\pdfoutput=1

\documentclass[11pt]{article}

\usepackage{acl}

\usepackage{times}
\usepackage{latexsym}

\usepackage{amsmath}
\usepackage{amssymb}
\usepackage{booktabs}
\usepackage[most]{tcolorbox} 
\usepackage{multirow}

\usepackage[utf8]{inputenc}
\usepackage[T1]{fontenc}
\usepackage{times}
\usepackage{latexsym}
\usepackage[margin=2.5cm]{geometry}
\usepackage{graphicx}
\usepackage{xcolor}
\usepackage{tcolorbox}
\usepackage{amsmath}
\usepackage{amssymb}
\usepackage{multicol}

\tcbuselibrary{skins, breakable}
\newcommand{\overarc}[1]{\overset{\frown}{#1}}

\definecolor{caseheader}{RGB}{129, 216, 208}
\definecolor{questionbg}{RGB}{240, 245, 252}
\definecolor{originalbg}{RGB}{255, 243, 235}
\definecolor{originalframe}{RGB}{210, 140, 90}
\definecolor{oursbg}{RGB}{235, 248, 240}
\definecolor{oursframe}{RGB}{60, 150, 100}
\definecolor{thinkcolor}{RGB}{80, 80, 160}
\definecolor{answercolor}{RGB}{180, 50, 50}
\definecolor{stepcolor}{RGB}{42, 100, 150}

\newcommand{\thinktag}{\textcolor{thinkcolor}{\texttt{<think>}}}
\newcommand{\thinktagclose}{\textcolor{thinkcolor}{\texttt{</think>}}}
\newcommand{\answertag}{\textcolor{answercolor}{\texttt{<answer>}}}
\newcommand{\answertagclose}{\textcolor{answercolor}{\texttt{</answer>}}}

\newtcolorbox{casestudy}[1][]{%
    enhanced,
    colback=white, colframe=caseheader,
    colbacktitle=caseheader, coltitle=white,
    fonttitle=\bfseries\sffamily,
    title={#1},
    top=6pt, bottom=6pt, left=8pt, right=8pt,
    boxrule=0.8pt, arc=2pt,
    toptitle=4pt, bottomtitle=4pt,
    drop shadow southeast={fill=black!15, opacity=0.4},
}

\newtcolorbox{questionbox}{%
    enhanced,
    colback=questionbg, colframe=caseheader!40,
    boxrule=0.5pt, arc=1.5pt,
    left=6pt, right=6pt, top=3pt, bottom=3pt,
    fontupper=\normalsize,
}

\newtcolorbox{originalbox}{%
    enhanced,
    colback=originalbg, colframe=originalframe,
    boxrule=0.5pt, arc=1.5pt,
    left=6pt, right=6pt, top=3pt, bottom=3pt,
    fontupper=\normalsize,
}

\newtcolorbox{oursbox}{%
    enhanced,
    colback=oursbg, colframe=oursframe,
    boxrule=0.5pt, arc=1.5pt,
    left=6pt, right=6pt, top=3pt, bottom=3pt,
    fontupper=\normalsize,   
}

\usepackage[T1]{fontenc}

\usepackage[utf8]{inputenc}

\usepackage{microtype}

\title{\textsc{OmniThoughtVis}: A Scalable Distillation Pipeline for Deployable Multimodal Reasoning Models}

\author{Yuanhao Yue, Chengyu Wang\thanks{\ \ \ Corresponding author.}, Yuanjie Lyu, Lei Shen, Jun Huang\\
Alibaba Group, Hangzhou, China\\
\texttt{\{yueyuanhao.yyh,chengyu.wcy,lyuyuanjie.lyj,}\\
\texttt{yuzhou.sl,huangjun.hj\}@alibaba-inc.com}\\
}

\begin{document}
\maketitle
\begin{abstract}
Recent multimodal large language models (MLLMs) have shown strong chain-of-thought (CoT) reasoning ability on vision-language tasks, but their direct deployment in real-world systems is often limited by latency and resource constraints. In practice, smaller MLLMs are preferred for online serving, yet their reasoning performance is bottlenecked by the lack of large-scale, high-quality multimodal CoT supervision.
In this paper, we present \textsc{OmniThoughtVis}, a scalable data curation and distillation pipeline for transferring multimodal reasoning capabilities from high-capacity teacher models to smaller, deployment-oriented MLLMs. Starting from a diverse open-source seed pool, our pipeline generates structured CoT traces and performs joint annotation of reasoning difficulty, answer quality, and semantic task tags. To maintain data quality at scale, we combine rule-based filtering, difficulty-aware selection, and tag-based diversity sampling, resulting in a curated corpus of 1.8M samples that supports controllable subset construction for downstream training.
We use \textsc{OmniThoughtVis} to distill Qwen3-VL models from 2B to 8B parameters and evaluate them on nine multimodal reasoning benchmarks. The resulting distilled models show consistent gains across model scales, including improvements of up to +16.8 points on MathVerse and +5.6 points on MMMU-Pro for the 4B model. Notably, the distilled 4B model matches or surpasses the undistilled 8B baseline on several tasks, highlighting the practical value of scalable reasoning distillation for deployment-oriented MLLMs.
\end{abstract}

\section{Introduction}

Recent multimodal large language models (MLLMs) have demonstrated strong chain-of-thought (CoT) reasoning ability on a wide range of vision-language tasks~\citep{bai2025qwen25vltechnicalreport}. However, directly deploying such models in real-world applications remains challenging: large models typically incur substantial latency, memory, and serving cost, while smaller models that are more suitable for online inference often underperform on reasoning-intensive tasks~\citep{hinton2015distillingknowledgeneuralnetwork,SurveyKD,wang2025distilqwen2}. This creates a practical gap between the reasoning quality of frontier MLLMs and the efficiency requirements of deployment-oriented systems.

A central bottleneck is the lack of large-scale, high-quality multimodal reasoning supervision. In the text domain, several large-scale CoT corpora have accelerated progress in natural language reasoning~\citep{10.5555/3722577.3722647,toshniwal2024openmathinstruct118millionmath,yu2024metamath,cai2025reasoning}. In contrast, open multimodal datasets are still dominated by instruction tuning or final-answer supervision, with limited support for structured, stepwise reasoning transfer~\citep{liu2023llava,ShareGPT4V,li2025llavaonevision}. As a result, smaller MLLMs fine-tuned on such data often exhibit shallow reasoning behavior and struggle on compositional or multi-step tasks. Moreover, naively generating multimodal CoT data at scale is insufficient in practice: synthetic traces can be noisy, redundant, and unevenly distributed across task types, reducing their usefulness for reliable distillation.

Recent efforts such as BEE~\citep{bee} and OpenMMReasoner~\citep{openmmreasoner} have advanced open multimodal data curation and reproducible training pipelines. Our work is complementary to these efforts, but focuses on a different practical objective: transferring reasoning ability into smaller, deployment-oriented MLLMs through a controllable large-scale distillation pipeline. In particular, rather than treating the source corpus as a ready-to-train dataset, we treat it as a seed pool and convert it into a distillation-oriented reasoning corpus by generating structured CoT traces, attaching joint annotations for reasoning difficulty, answer quality, and semantic task tags, and applying filtering and diversity-aware subset selection for downstream training.

To this end, we present \textsc{OmniThoughtVis}, a scalable data curation and distillation pipeline for multimodal reasoning. Starting from a diverse open-source seed pool, our pipeline distills structured CoT traces from a high-capacity teacher model under explicit format constraints. We then perform joint annotation of reasoning difficulty, answer quality, and semantic task tags, followed by rule-based filtering, difficulty-aware selection, and tag-based diversity sampling. This process yields a curated corpus of 1.8M samples and supports controllable subset construction for training smaller reasoning-capable MLLMs under practical compute and deployment constraints.

We evaluate distilled Qwen3-VL models ranging from 2B to 8B parameters on nine multimodal reasoning benchmarks. Distillation with \textsc{OmniThoughtVis} leads to consistent improvements across model scales, including gains of up to +16.8 points on MathVerse and +5.6 points on MMMU-Pro for the 4B model. Notably, the distilled 4B model matches or surpasses the undistilled 8B baseline on several tasks, suggesting a favorable quality-efficiency trade-off for deployment-oriented multimodal reasoning systems. To support reproducible research and practical development, we will release the curated dataset, data pipeline, and model checkpoints.

Our main contributions are as follows:
\begin{itemize}
\item We present \textsc{OmniThoughtVis}, a scalable pipeline for producing multimodal CoT supervision tailored to distilling reasoning capabilities into smaller, deployment-oriented MLLMs.
\item We introduce a practical data curation recipe that combines structured teacher generation with joint annotation, difficulty-aware filtering, and diversity-aware sampling, enabling controllable reasoning data construction at million-sample scale.
\item We show that \textsc{OmniThoughtVis} consistently improves distilled 2B--8B models across nine benchmarks, and we report practical observations on data selection and scaling behavior that are useful for building efficient multimodal reasoning systems.
\end{itemize}

\section{Related Work}

\noindent\textbf{Models with Extended Reasoning.}
Chain-of-thought (CoT) prompting~\citep{cot} has shown that exposing intermediate reasoning steps can substantially improve model performance on complex tasks. Follow-up work has explored transferring such reasoning behavior into smaller models using teacher-generated rationales or step-by-step supervision~\citep{hsieh-etal-2023-distilling,ho-etal-2023-large,mukherjee2023orca}. More recently, reinforcement learning with verifiable rewards has further strengthened long-form reasoning in frontier models, as illustrated by DeepSeek-R1~\citep{DeepSeek-R1} and OpenAI o1. In parallel, the community has begun to curate large-scale reasoning datasets for open models~\citep{openthoughts,cai2025reasoning}. Our work is complementary to these efforts: rather than proposing a new reasoning algorithm, we focus on the practical problem of producing scalable multimodal reasoning supervision for distilling deployment-oriented MLLMs.

\noindent\textbf{Multimodal Dataset Curation and Reasoning Supervision.}
A large body of prior work has improved multimodal models through instruction tuning and synthetic data generation. LLaVA~\citep{liu2023llava} introduced visual instruction tuning with GPT-generated conversations, but its supervision is primarily short-form and does not provide structured reasoning traces. ScienceQA~\citep{scienceQA} includes CoT-style explanations, but its scale is limited and its domain is restricted to science questions. ShareGPT4V~\citep{ShareGPT4V} scales up visual description data using GPT-4V, yet mainly targets perceptual grounding rather than multi-step reasoning. Cambrian-1~\citep{tong2024cambrian} aggregates diverse vision-language resources for broad instruction tuning, but does not explicitly optimize for reasoning-oriented distillation into smaller models.

More recent open efforts such as BEE~\citep{bee} and OpenMMReasoner~\citep{openmmreasoner} have pushed multimodal data curation and reproducible training toward more capable reasoning models. Our work differs in emphasis. Rather than treating large open corpora as directly usable training data, we frame them as seed pools for building a distillation-oriented reasoning corpus under practical constraints. \textsc{OmniThoughtVis} adds three ingredients that are central to this goal: (1) structured CoT traces produced under explicit output constraints, (2) joint annotations of reasoning difficulty, answer quality, and semantic task tags, and (3) filtering and diversity-aware subset selection designed to improve reasoning transfer to smaller models. This combination enables finer control over training data composition and supports practical multimodal reasoning distillation at million-sample scale.

\begin{figure}[t]
    \centering
    \includegraphics[width=.875\linewidth]{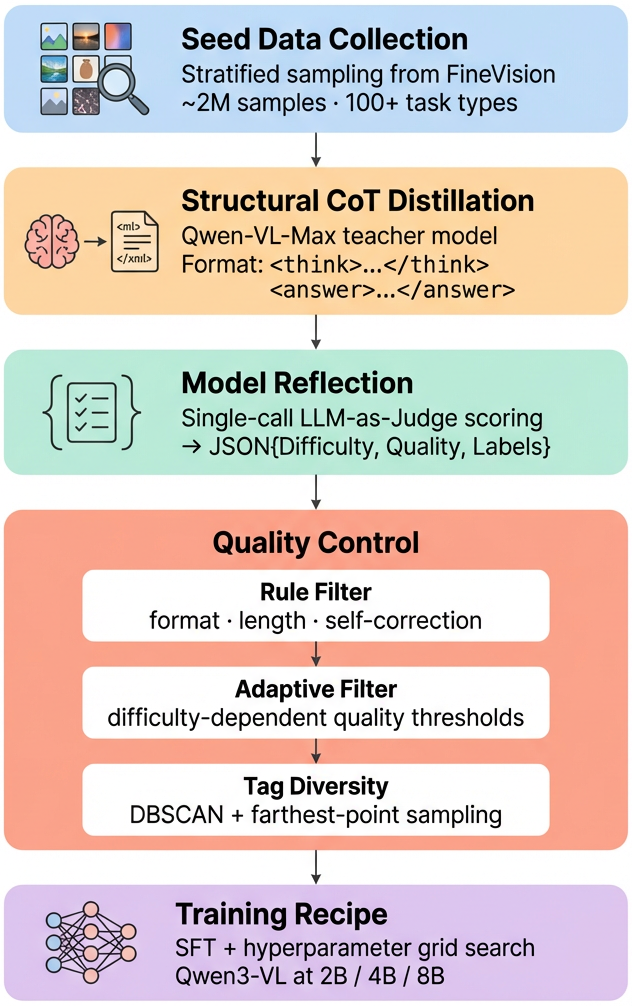}
    \caption{Overview of the \textsc{OmniThoughtVis} data curation and distillation pipeline. Starting from a broad open-source seed pool, we generate structured multimodal CoT traces, apply joint annotation and quality control, and construct training subsets for distilling smaller reasoning-capable MLLMs.}
    \label{fig:pipeline}
\end{figure}

\section{Methodology}
\label{sec:methodology}

We next describe the \textsc{OmniThoughtVis} pipeline for building a distillation-oriented multimodal reasoning corpus from open-source seed data. Figure~\ref{fig:pipeline} provides an overview. Our goal is not simply to aggregate more multimodal samples, but to transform a broad seed pool into a curated corpus that supports controllable reasoning transfer to smaller, deployment-oriented MLLMs.

\subsection{Seed Data Sampling}
\label{sec:seed-sampling}

We start from FineVision~\citep{finevision}, a large-scale aggregation of publicly available open-source datasets spanning visual question answering, chart understanding, mathematical reasoning, and general visual reasoning. In our pipeline, FineVision serves as a \emph{seed pool} rather than a ready-to-train reasoning dataset. We sample from it using stratified sampling to preserve domain and task diversity, while capping each category at 20K samples to prevent a small number of high-frequency sources from dominating the candidate pool. This procedure yields an initial pool of 3.5M multimodal samples covering a broad range of domains and task types.

\subsection{Distillation with Format Constraints}
\label{sec:structured-distillation}

We use Qwen-VL-Max~\citep{qwen3vl} as the teacher model to generate CoT traces for retained samples. Decoding is performed with temperature $T=0.5$, which we found to provide a practical balance between trace diversity and output stability, and the maximum generation length is set to 8192 tokens. To make the outputs easier to parse, score, and filter at scale, we enforce a structured XML-style format that explicitly separates the reasoning trace from the final answer, as shown below. Samples that do not satisfy the required format are removed during subsequent quality control. Detailed prompts are provided in Appendix~\ref{sec:prompt}.

\begin{tcolorbox}[
  colback=gray!5, colframe=gray!60,
  boxrule=0.4pt, arc=2pt,
  left=6pt, right=6pt, top=4pt, bottom=4pt,
  fontupper=\small
]
\texttt{<think>} \textit{Step-by-step reasoning trace} \texttt{</think>} \\
\texttt{<answer>} \textit{Final answer} \texttt{</answer>}
\end{tcolorbox}

This explicit structure serves two practical purposes. First, it enables robust downstream parsing for large-scale automated curation. Second, it supports separate handling of intermediate reasoning and final predictions during filtering, evaluation, and future reuse of the dataset.

\subsection{Joint Scoring and Semantic Tagging}
\label{sec:scoring-system}

A central component of our pipeline is the \emph{joint} annotation of each generated sample along three axes: reasoning difficulty, answer quality, and semantic task tags. Given an image, instruction, and teacher response, we prompt a scorer model to output a JSON object containing these fields.\footnote{See Appendices~\ref{sec:score} and~\ref{sec:tag} for implementation details.} Joint annotation reduces redundant inference passes and helps maintain consistency across annotations at million-sample scale.

\noindent\textbf{Difficulty (1--5):} This score reflects the cognitive complexity of the underlying task. Level~1 corresponds to simple recognition such as object presence, color, or shape identification. Level~2 covers basic counting or spatial relations. Level~3 involves moderate reasoning over actions or attributes. Level~4 denotes more challenging multi-step reasoning involving subtle visual cues or uncommon concepts. Level~5 represents abstract reasoning, complex scene understanding, or ambiguous contexts.

\noindent\textbf{Quality (1--5):} This score reflects the estimated correctness and completeness of the generated response. Level~1 denotes a fully incorrect or irrelevant answer, Level~3 indicates partial correctness, and Level~5 denotes a response judged to be accurate and complete.

\begin{figure}[t]
    \centering
    \includegraphics[width=0.99\linewidth]{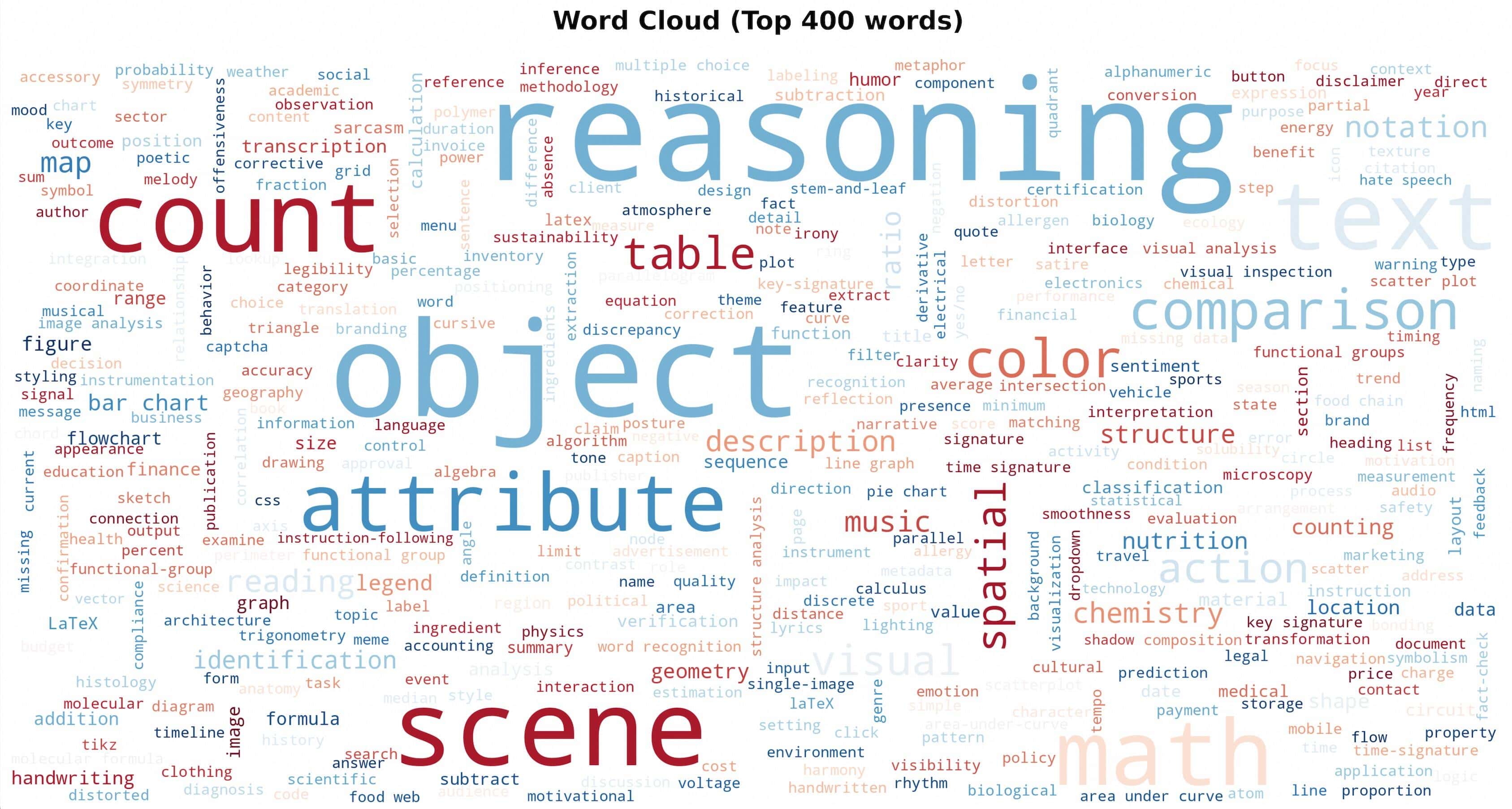}
    \caption{Word cloud visualization of the top 400 task-related labels in \textsc{OmniThoughtVis}. Frequent labels such as \textit{reasoning}, \textit{comparison}, \textit{count}, \textit{object}, and \textit{scene} suggest broad semantic coverage across visual understanding and multimodal reasoning tasks.}
    \label{fig:wordcloud}
\end{figure}

\noindent\textbf{Semantic tags:} We additionally generate a set of task tags (e.g., \texttt{counting}, \texttt{spatial}, \texttt{reasoning}, \texttt{math}, \texttt{object}) to characterize the sample. The scorer outputs only a JSON object, enabling robust automated parsing and downstream selection. Frequent tags are visualized in Figure~\ref{fig:wordcloud}.

We use Qwen3-VL-Flash as the scorer because annotation throughput is a practical bottleneck in million-sample curation, and this model offers a favorable trade-off between multimodal understanding quality and inference efficiency. This LLM-as-a-judge setup~\citep{Chatbot-Arena} provides a scalable way to attach metadata to generated traces. In our experiments, difficulty proved to be the most actionable signal for downstream data selection, while quality scores were more useful as an auxiliary inspection signal than as a strong standalone filter.

\subsection{Quality Control and Subset Construction}
\label{sec:quality-control}

We apply quality control in three stages, followed by subset construction for training.

\noindent\textbf{Stage 1 (Rule-Based Filtering):} We remove samples that (i) lack the required output tags, (ii) have CoT traces shorter than 20 tokens or longer than 4000 tokens, (iii) contain placeholder text or strong repetition, or (iv) exhibit unstable generation patterns such as repeated restarts, unresolved contradictions, or abrupt self-corrections. We note that self-correction is not always undesirable; however, in our pipeline, these patterns often correlate with malformed or low-confidence traces that are difficult to parse and verify automatically. The 4000-token upper bound is used as a practical trade-off between preserving rich supervision and avoiding excessively long traces that often contain repetition or drift while substantially increasing training cost. Approximately 5\% of samples are removed at this stage.

\noindent\textbf{Stage 2 (Difficulty-Aware Selection):} Rather than applying a fixed quality-only threshold, we compare multiple filtering strategies in Section~\ref{sec:analysis}. Based on these experiments, we adopt Difficulty~$\geq$~4 as the primary data selection criterion for the training subset. Empirically, this criterion retains more reasoning-intensive samples and yields better transfer than filtering by quality alone.

\noindent\textbf{Stage 3 (Tag-Based Diversity Sampling):} To improve coverage beyond dominant task categories, we perform diversity-aware subset selection using semantic tags. We embed tags using a text embedding model (e.g., \texttt{Qwen3-Embedding}~\citep{qwen3-embedding}) and cluster them with DBSCAN~\citep{DBSCAN} to merge near-synonymous tags (e.g., \texttt{spatial} and \texttt{position}). Each sample is then represented by the average of its clustered tag embeddings, producing a coarse reasoning-profile vector. We apply farthest-point sampling in this space to construct a more diverse subset and reduce over-representation of frequent categories such as object-centric recognition. This tag space also enables domain-targeted curation: practitioners can filter or resample by tags (e.g., spatial reasoning or chart interpretation) without regenerating the full corpus.

\noindent\textbf{Training subset construction:} The full \textsc{OmniThoughtVis} corpus contains 1.8M curated samples. In this paper, we train on selected subsets under fixed compute budgets to enable controlled comparison across model scales. This setup reflects a practical deployment-oriented workflow in which a large candidate pool is curated once, and task- or budget-specific subsets are then selected for downstream training.

\begin{figure}[t]
    \centering
    \includegraphics[width=0.73\linewidth]{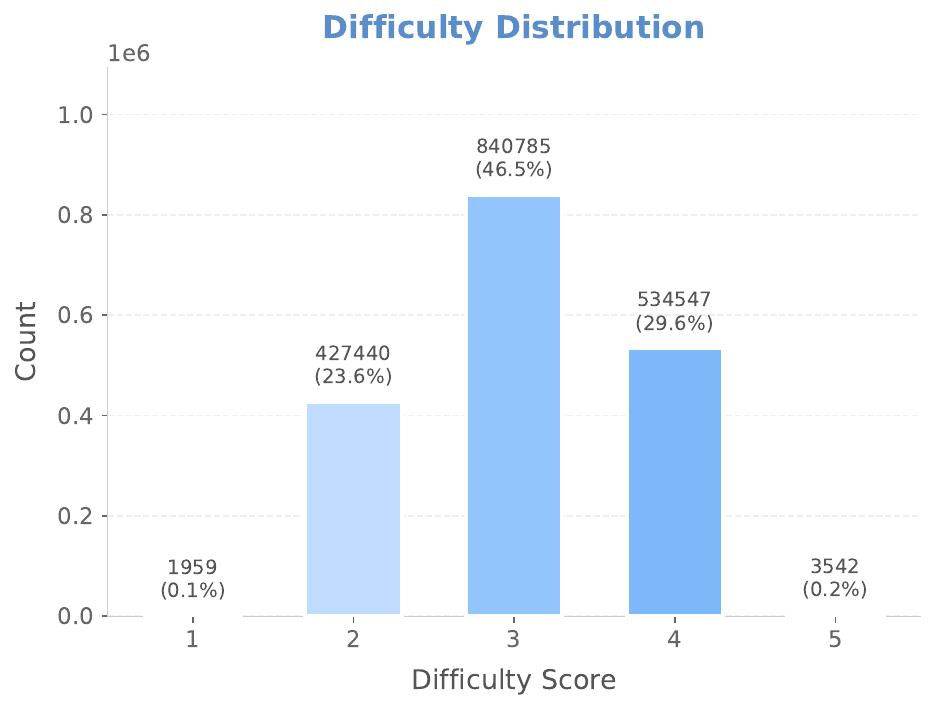}\\[0.5em]
    \includegraphics[width=0.73\linewidth]{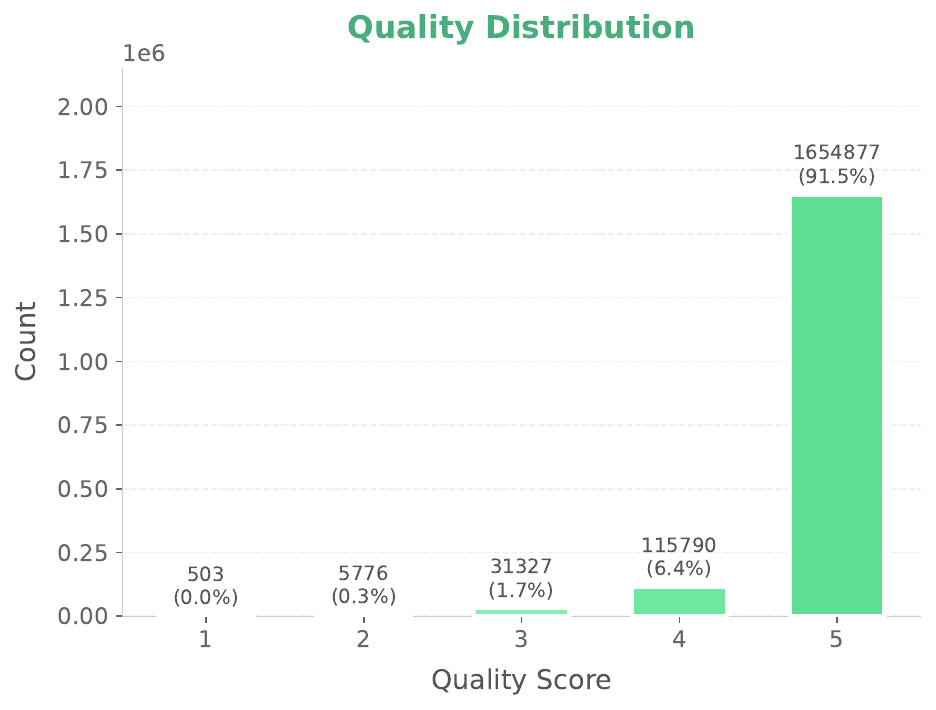}
    \caption{Difficulty (top) and quality (bottom) distributions in \textsc{OmniThoughtVis}. Difficulty spans a broad range, while quality is concentrated in the high-score regime, reflecting the generally strong teacher-generated responses in the retained corpus.}
    \label{fig:dataset-stats}
\end{figure}

\subsection{Dataset Statistics}
\label{sec:dataset-stats}

Figure~\ref{fig:dataset-stats} summarizes the key statistics of \textsc{OmniThoughtVis}. The difficulty distribution is approximately unimodal and centered at Level~3 (46.5\%), with Levels~2 (23.6\%) and 4 (29.6\%) also well represented; extreme levels are rare (Level~1: 0.1\%, Level~5: 0.2\%). The quality distribution is strongly skewed toward high ratings, with 91.5\% of samples assigned Level~5 and 6.4\% assigned Level~4. This concentration suggests that quality scores are less discriminative than difficulty scores in our current setup, which is consistent with our later finding that difficulty is the more useful signal for training subset selection. Analysis of semantic tags confirms broad domain coverage, including \texttt{reasoning} (85\%), \texttt{object} (75\%), \texttt{scene} (45\%), \texttt{count} (38\%), \texttt{text} (37\%), and \texttt{math} (33\%). Overall, \textsc{OmniThoughtVis} is characterized not only by scale, but also by structured reasoning supervision and controllable metadata for downstream curation.

\begin{table*}[t]
\centering
\footnotesize
\setlength{\tabcolsep}{3.5pt}
\renewcommand{\arraystretch}{0.92}
\begin{tabular}{lccccccccc}
\toprule
\multirow{2}{*}{\textbf{Model}} & \multicolumn{2}{c}{\textbf{General}} & \multicolumn{4}{c}{\textbf{Knowledge \& Understanding}} & \multicolumn{3}{c}{\textbf{Mathematical Reasoning}} \\
\cmidrule(lr){2-3} \cmidrule(lr){4-7} \cmidrule(lr){8-10}
 & MMB & MMStar & AI2D & MMMU\textsubscript{val} & MMMU\textsubscript{Pro-S} & MMMU\textsubscript{Pro-V} & MathVista & MathVerse & MathVision \\
\midrule
Qwen3-VL-2B            & 74.3 & 56.6 & 71.3 & 47.9 & 32.1 & 27.5 & 60.0 & 34.1 & 20.4 \\
DistilQwen3-VL-2B      & \textbf{74.7} & \textbf{61.5} & \textbf{75.4} & \textbf{49.2} & \textbf{35.6} & \textbf{31.9} & \textbf{65.0} & \textbf{43.0} & \textbf{27.3} \\
\midrule
Qwen3-VL-4B            & 82.0 & 66.6 & 79.4 & 56.7 & 41.3 & 40.3 & 70.7 & 36.4 & 31.9 \\
DistilQwen3-VL-4B      & \textbf{82.7} & \textbf{68.4} & \textbf{81.6} & \textbf{60.3} & \textbf{46.9} & \textbf{45.2} & \textbf{74.0} & \textbf{53.2} & \textbf{32.9} \\
\midrule
Qwen3-VL-8B            & 82.6 & 69.3 & 81.0 & 60.0 & 46.9 & 45.3 & 72.5 & 38.2 & 35.2 \\
DistilQwen3-VL-8B      & \textbf{84.7} & \textbf{71.1} & \textbf{83.7} & \textbf{62.8} & \textbf{50.1} & \textbf{48.6} & \textbf{74.8} & \textbf{55.5} & \textbf{40.1} \\
\bottomrule
\end{tabular}
\caption{Main results comparing Qwen3-VL baseline models and models fine-tuned on \textsc{OmniThoughtVis}. MMB = MMBench, Pro-S = Pro\textsubscript{Standard}, and Pro-V = Pro\textsubscript{Vision}. All values are percentages (\%).}
\label{tab:main-results}
\end{table*}

\section{Experiments}
\label{sec:experiments}

We conduct experiments to evaluate whether \textsc{OmniThoughtVis} improves reasoning transfer to smaller multimodal models under a practical distillation setup. Our analysis focuses on three questions: (1) whether the curated corpus improves model performance across scales, (2) which data selection choices matter most in practice, and (3) how reasoning performance evolves during training.

\subsection{Experimental Setup}
\label{sec:exp-setup}

\noindent\textbf{Training Protocol.}
We initialize three models from the Qwen3-VL family~\citep{qwen3vl} as student backbones: Qwen3-VL-2B, Qwen3-VL-4B, and Qwen3-VL-8B. We refer to the resulting distilled models as the DistilQwen3-VL series. Unless otherwise specified, all models are trained on a selected 0.5M-sample subset from the full 1.8M-sample \textsc{OmniThoughtVis} corpus. We use a fixed 0.5M subset to enable controlled comparison across model scales under a common compute budget; the larger corpus remains important as a candidate pool for difficulty-aware and tag-aware subset construction.

Training uses AdamW ($\beta_1 = 0.9$, $\beta_2 = 0.999$, weight decay $= 0.01$), batch size 128, sequence length 3072, linear warmup over 3\% of training steps, and cosine learning rate decay to 10\% of the peak value. The original Qwen3-VL-Instruct models serve as baselines so that observed gains can be attributed to fine-tuning on \textsc{OmniThoughtVis} rather than changes in backbone architecture.

We perform a grid search over learning rates $\{5{\times}10^{-6}, 1{\times}10^{-5}, 2{\times}10^{-5}, 5{\times}10^{-5}\}$ and training epochs $\{1,2,3,4,5\}$ on the 2B model, evaluating all 20 configurations on a held-out 10K-sample split. Figure~\ref{fig:hp-search} shows that the best setting uses a learning rate of $5{\times}10^{-6}$ for 4 epochs, which we adopt for all model scales. Higher learning rates generally lead to weaker performance, while additional epochs beyond the best setting do not provide consistent gains across learning rates.

\noindent\textbf{Benchmarks.}
We evaluate on nine multimodal reasoning benchmarks. For \textit{general visual understanding and knowledge-intensive reasoning}, we use AI2D~\citep{ai2d}, MMBench~\citep{mmbench}, MMMU~\citep{mmmu}, MMMU-Pro (Standard and Vision)~\citep{mmmupro}, and MMStar~\citep{mmstar}. For \textit{mathematical reasoning}, we evaluate on MathVista~\citep{mathvista}, MathVerse~\citep{mathverse}, and MathVision~\citep{mathvision}. Together, these benchmarks cover tasks ranging from visual comprehension to multi-step multimodal inference.

\noindent\textbf{Evaluation Protocol.}
All models are evaluated with thinking prompts (Appendix~\ref{sec:prompt}) and decoding temperature 0.5. For multiple-choice benchmarks, we report exact-match accuracy after extracting predictions from \texttt{<answer>} tags. For open-ended tasks requiring LLM-based judging, we use Qwen3-VL-Plus. Because teacher generation, automatic scoring, and student backbones all involve Qwen-family models, some family-specific bias may remain; we therefore place primary emphasis on objective benchmark results and treat judge-based evaluation as a secondary signal.

\subsection{Main Results}
\label{sec:main-results}

Table~\ref{tab:main-results} compares baseline Qwen3-VL models with their distilled counterparts trained on \textsc{OmniThoughtVis}. Distillation yields consistent gains across all three model scales. For the 4B model, improvements are especially pronounced on reasoning-intensive benchmarks such as MathVerse (+16.8) and MMMU-Pro (+5.6), suggesting that structured multimodal CoT supervision can effectively transfer reasoning behavior to smaller models. The 2B model shows similar gains, with MathVerse improving from 34.1 to 43.0 and MathVision from 20.4 to 27.3. The 8B model achieves the strongest absolute scores, reaching 83.7 on AI2D and 55.5 on MathVerse.

Notably, the distilled 4B model matches or surpasses the undistilled 8B baseline on several benchmarks (e.g., 81.6 vs.\ 81.0 on AI2D and 60.3 vs.\ 60.0 on MMMU). While these comparisons do not replace direct deployment measurements, they indicate a favorable quality--efficiency trade-off and suggest that scalable reasoning distillation can improve the usefulness of smaller models in deployment-oriented settings.

\subsection{Detailed Analysis}
\label{sec:ablation}

\begin{figure}[t]
    \centering
    \includegraphics[width=0.98\linewidth]{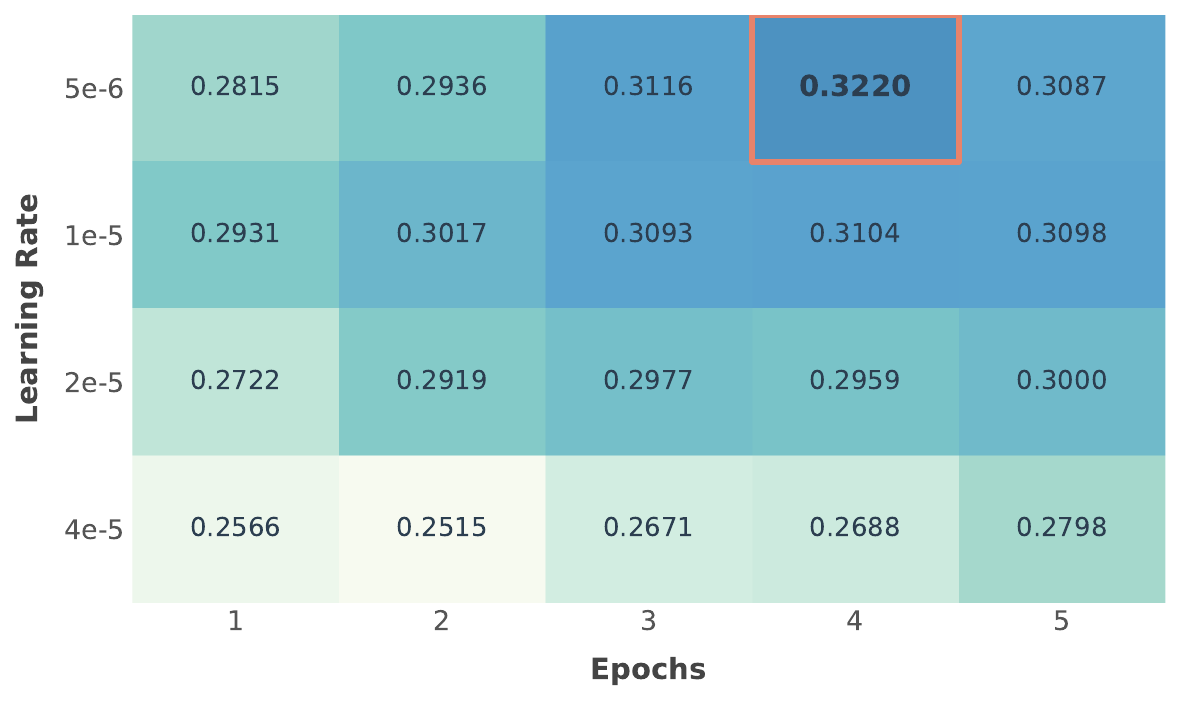}
    \caption{Hyperparameter search on MMMU\textsubscript{Pro-V} for DistilQwen3-VL-2B. Baseline: 0.2130.}
    \label{fig:hp-search}
\end{figure}

\noindent\textbf{Hyperparameter Search.}
We perform a grid search over learning rates and training epochs on the validation set. For DistilQwen3-VL-2B, we evaluate four learning rates ($5 \times 10^{-6}, 1 \times 10^{-5}, 2 \times 10^{-5}, 5 \times 10^{-5}$) across 1--5 epochs, using MMMU-Pro Vision as the selection metric. As shown in Figure~\ref{fig:hp-search}, hyperparameter choice has a substantial effect. The best configuration uses a learning rate of $5 \times 10^{-6}$ for 4 epochs, achieving a score of 0.3220 compared with the 0.2130 baseline. Higher learning rates generally degrade performance, and the best performance is reached before the final epoch for the strongest low-learning-rate setting. We therefore use 4 epochs as the default setting in subsequent experiments.

\noindent\textbf{Data Filtering Strategies.}
We compare filtering by quality and/or difficulty under otherwise identical settings, with each model trained on approximately 100K samples (Table~\ref{tab:filter-ablation}). Quality-only filtering (Quality $\geq$ 5) underperforms random sampling (0.2948 vs.\ 0.3093), whereas difficulty-only filtering (Difficulty $\geq$ 4) performs best (0.3156). Combining both criteria (Quality $\geq$ 5 \& Difficulty $\geq$ 4) also improves over random sampling (0.3139), but remains slightly weaker than difficulty-only selection. These results suggest that difficulty is the more useful selection signal in our current pipeline, while quality scores are less discriminative for subset construction.

\begin{figure}[t]
    \centering
    \includegraphics[width=0.88\linewidth]{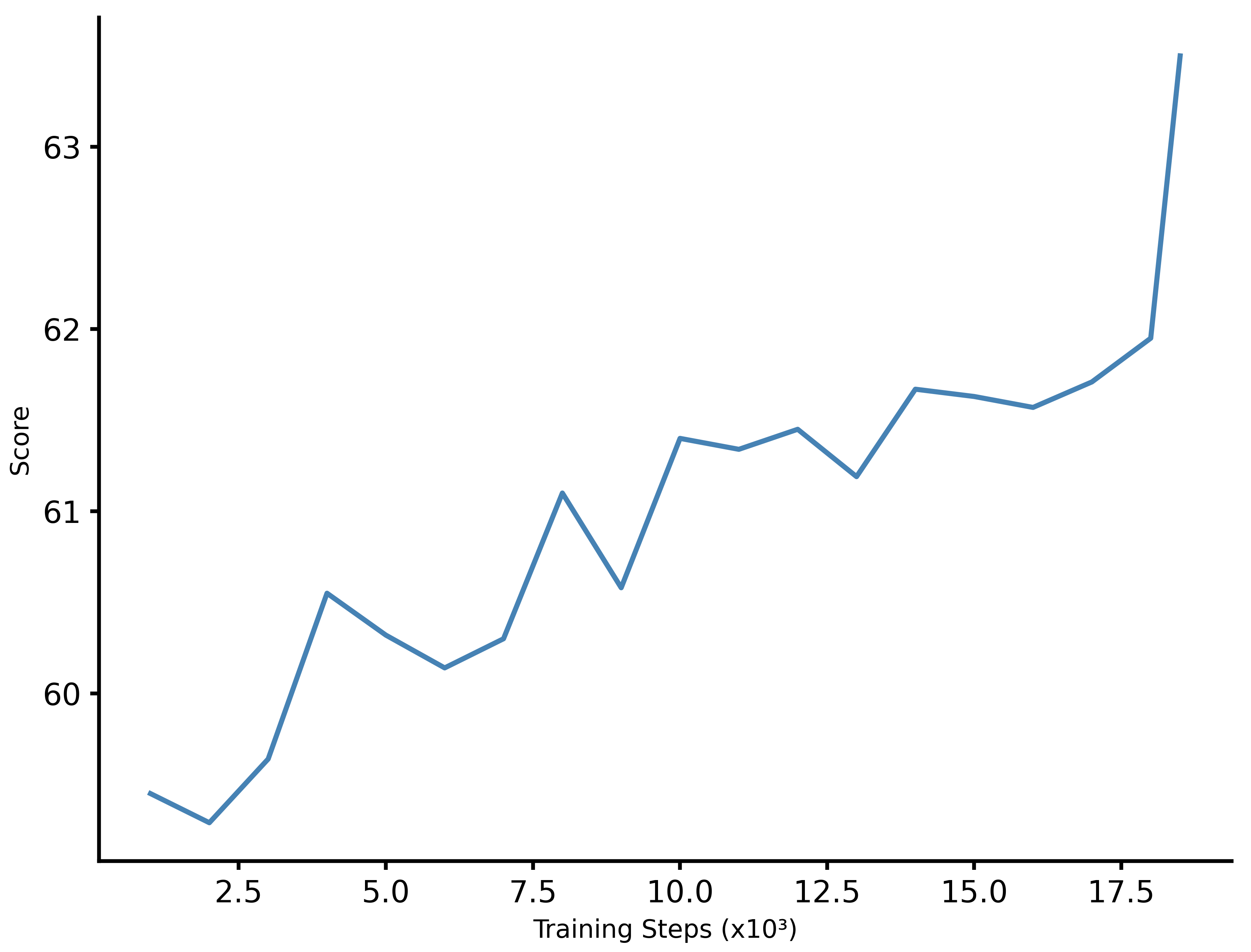}
    \caption{Average performance across nine benchmarks during DistilQwen3-VL-8B training on \textsc{OmniThoughtVis}. Performance improves from 59.5 to 63.5 over 19K steps.}
    \label{fig:avg-scaling}
\end{figure}

\begin{figure}[t]
    \centering
    \includegraphics[width=\linewidth]{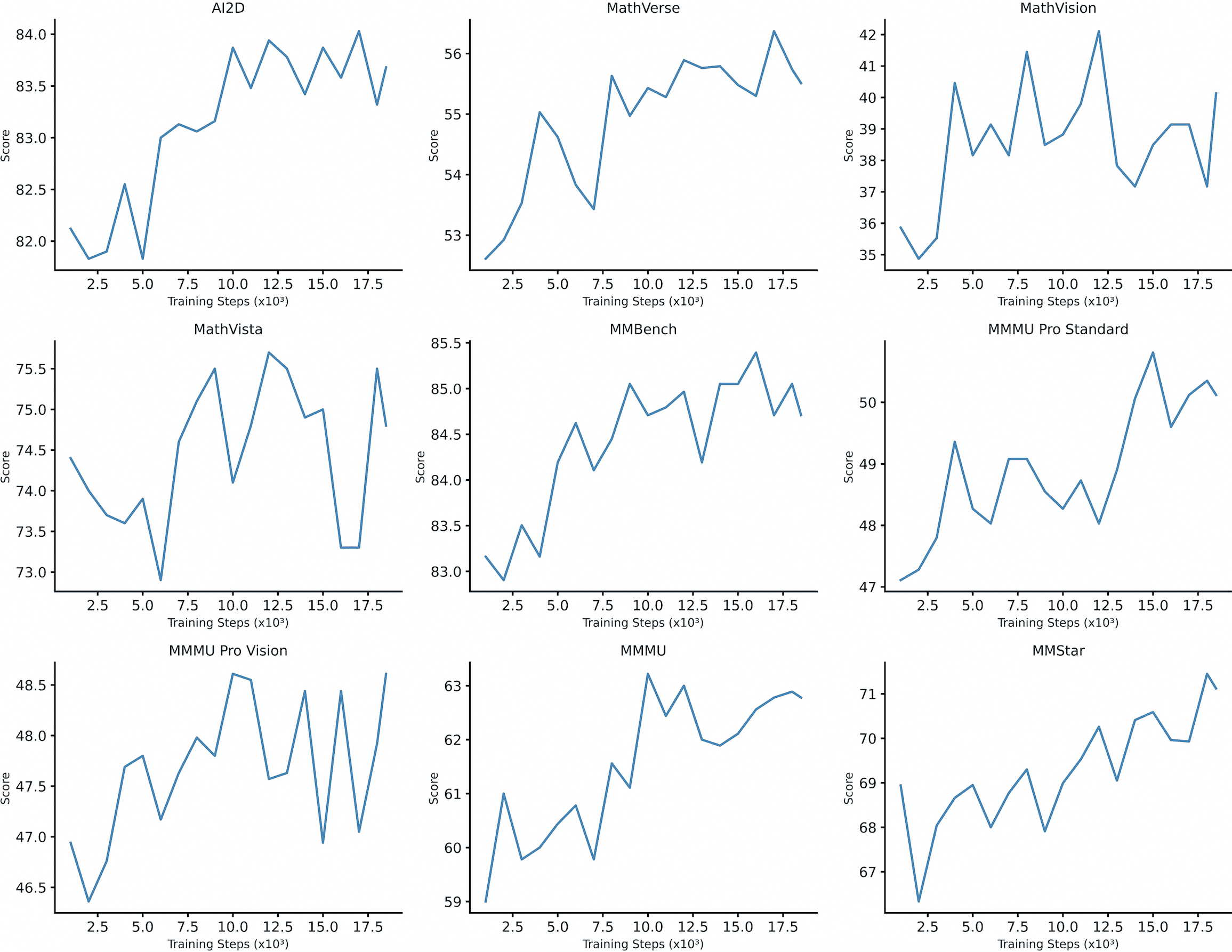}
    \caption{Per-benchmark scaling for DistilQwen3-VL-8B. General visual benchmarks plateau earlier, while reasoning-intensive benchmarks continue to improve.}
    \label{fig:benchmark-scaling}
\end{figure}

\noindent\textbf{Scaling Behavior During Training.}
We further analyze training-time scaling by evaluating DistilQwen3-VL-8B across all nine benchmarks. Figure~\ref{fig:avg-scaling} shows the average performance trajectory, while Figure~\ref{fig:benchmark-scaling} reports per-benchmark trends. The average score increases from approximately 59.5 at 1K steps to 63.5 at 19K steps, indicating continued gains throughout training. These gains are not uniform across tasks: general visual understanding benchmarks such as AI2D and MMBench improve quickly and then plateau, whereas reasoning-intensive benchmarks such as MathVerse, MMMU, and MMMU-Pro continue to improve for longer.

This difference suggests that general understanding and reasoning transfer may saturate on different timescales. In practical terms, it indicates that uniform training schedules may not be optimal for multimodal reasoning distillation, and motivates future curriculum or stage-wise training strategies that allocate more training emphasis to reasoning-heavy data later in training.

\begin{table}[t]
\centering
\begin{tabular}{lc}
\toprule
\textbf{Filtering Strategy} & \textbf{Score} \\
\midrule
No filtering (random 100K) & 0.3093 \\
Quality $\geq$ 5 only & 0.2948 \\
Difficulty $\geq$ 4 only & 0.3156 \\
Quality $\geq$ 5 \textbf{and} Difficulty $\geq$ 4 & 0.3139 \\
\bottomrule
\end{tabular}
\caption{Ablation of data filtering strategies. All models are trained on approximately 100K samples.}
\label{tab:filter-ablation}
\end{table}

\subsection{Practical Insights}
\label{sec:analysis}

Our experiments suggest three practical lessons for large-scale multimodal reasoning distillation.

\noindent\textbf{Difficulty is a more actionable data-selection signal than quality.}
In our current setup, quality scores are concentrated in the high-score regime and are therefore less useful for separating informative from uninformative training examples. By contrast, selecting more difficult samples consistently yields stronger reasoning transfer. This makes difficulty a more practical signal for subset construction in large-scale synthetic reasoning pipelines.

\noindent\textbf{Reasoning transfer and general understanding do not saturate at the same rate.}
Scaling curves show that general visual understanding often improves early and then plateaus, while reasoning-heavy benchmarks continue to benefit from additional training. This suggests that multimodal distillation pipelines may benefit from stage-wise or curriculum-based scheduling rather than uniform training across all examples.

\noindent\textbf{A large curated pool is useful even when the final training subset is smaller.}
Although our main experiments train on 0.5M selected samples, the full 1.8M-sample corpus remains valuable because it provides enough coverage for filtering, diversity-aware resampling, and future task-specific subset construction. In practical settings, curating a large candidate pool once and selecting smaller training subsets later can be more flexible than repeatedly regenerating task-specific data.

Due to space limitations, additional case studies are provided in Appendix~\ref{case_study}.


\section{Conclusion}

We presented \textsc{OmniThoughtVis}, a scalable data curation and distillation pipeline for transferring multimodal reasoning capabilities into smaller, deployment-oriented MLLMs. Starting from a broad open-source seed pool, our pipeline combines structured CoT generation, joint annotation, and downstream subset selection to construct a 1.8M-sample reasoning corpus with controllable metadata. Our experiments show that this pipeline consistently improves distilled models across nine benchmarks, with especially strong gains on reasoning-intensive tasks and a favorable quality--efficiency trade-off for smaller models.

Beyond the released corpus itself, we view \textsc{OmniThoughtVis} as a practical recipe for large-scale multimodal reasoning distillation. In particular, our results suggest that difficulty-aware selection is more useful than quality-only filtering in the current setup, and that reasoning-heavy benchmarks may benefit from different training dynamics than general visual understanding tasks. We will release the curated dataset, pipeline, and model checkpoints to support reproducible research and the practical development of efficient multimodal reasoning systems.

\section*{Limitations}

While \textsc{OmniThoughtVis} covers a broad range of multimodal reasoning tasks and yields consistent gains in our experiments, several limitations remain.
First, our current pipeline primarily focuses on English-language, vision-centric data. Extending the approach to additional languages and modalities would improve its generalizability.
Second, the dataset is constructed through automated teacher generation and model-based scoring. Although this enables scalable curation, some generated reasoning traces or metadata may still contain noise or model-specific biases. Further validation with broader evaluation settings would strengthen confidence in the resulting corpus.
Third, our experiments focus on benchmark-based evaluation under a fixed training budget. Additional studies on more diverse deployment scenarios and larger-scale training configurations would help further assess the practical utility of the pipeline.
We leave these directions to future work.

\section*{Ethical Considerations}

\textsc{OmniThoughtVis} is constructed from publicly available, open-source vision-language datasets using automated generation, scoring, and filtering procedures. Although these steps are intended to improve data quality and task diversity, several ethical considerations remain.
First, the dataset may inherit biases present in the source data or teacher models, which could affect the behavior of distilled models. Second, because the annotations are largely automated, some samples may contain inaccurate reasoning traces or metadata. Third, the dataset is primarily English-language and vision-centric, which may limit coverage across languages, domains, and populations.
All source datasets were checked for license compatibility to the best of our knowledge, and we only include data intended for research use. We encourage responsible use and additional evaluation before applying models trained on \textsc{OmniThoughtVis} in downstream settings.


\appendix

\section{Technical Details}

\subsection{Distillation Prompt Template}
\label{sec:prompt}

Below, we detail the standard prompt template used when collecting long-form CoT traces from high-capacity multi-modal teacher models:
\paragraph{System Prompt.}
The system prompt to the teacher is as follows:
\begin{quote}
\texttt{You are a helpful assistant to think step by step. Provide your reasoning steps within <think></think> tags and give your final answer within <answer></answer> tags.}
\end{quote}

\paragraph{Query Structure.}
Given an image and a natural language question, the input to the teacher follows the structured format:
\begin{quote}
\texttt{\#\#\# Image}\\
\texttt{<image>}\\
\texttt{\#\#\# Question}\\
\texttt{\{question\}}\\
\texttt{\#\#\# Output Format (Strictly Enforced)}\\
\texttt{<think>}\\
\texttt{Clearly explain your reasoning step by step. Describe how you arrived at the conclusion.
The reasoning process MUST BE enclosed within <think> </think> tags.}\\
\texttt{/<think>}\\
\texttt{<answer>}\\
\texttt{Your final answer to the user's question.}\\
\texttt{</answer>}\\
\end{quote}

\subsection{Automatic Scoring Protocol}
\label{sec:score}

For data selection and curation, we use Qwen3-VL-Flash as an automatic scorer to evaluate each sample along two axes: reasoning difficulty and annotation quality. Each sample is also assigned a set of open-ended task tags; here we focus on the numeric scoring protocol.

\paragraph{Reasoning Difficulty.}
Each sample is assigned a difficulty score $d_i \in \{1,2,3,4,5\}$ based on the complexity of the underlying vision-language reasoning required. The criteria are:
\begin{itemize}
    \item \textbf{1 (Very Easy):} Object is plainly present; requires simple color or shape identification.
    \item \textbf{2 (Easy):} Involves basic counting of clearly visible items, or straightforward spatial relationships.
    \item \textbf{3 (Moderate):} Requires brief reasoning or identification of common actions or attributes.
    \item \textbf{4 (Hard):} Demands multi-step reasoning, attention to subtle visual cues, or handling rare concepts.
    \item \textbf{5 (Very Hard):} Involves abstract reasoning, complex scene understanding, or highly ambiguous context.
\end{itemize}

\paragraph{Annotation Quality.}
Each annotated output is assigned a quality score $q_i \in \{1,2,3,4,5\}$ reflecting estimated correctness, relevance, and completeness, defined as follows:
\begin{itemize}
    \item \textbf{1 (Very Low):} Entirely incorrect or irrelevant answer.
    \item \textbf{2 (Low):} Mostly incorrect, with only minor correct elements present.
    \item \textbf{3 (Medium):} Partially correct; key details are missing or incorrect elements are present.
    \item \textbf{4 (High):} Largely correct, with only minor inaccuracies or omissions.
    \item \textbf{5 (Very High):} Fully accurate, precise, and complete response.
\end{itemize}

These scoring guidelines provide a practical framework for large-scale multimodal data curation. In the main experiments, we find that difficulty is the more useful signal for subset selection, while quality is primarily used as an auxiliary indicator during inspection and filtering.

\subsection{Semantic Tag Definitions and Coverage}
\label{sec:tag}

To support fine-grained subset construction and downstream analysis, each sample in the dataset is assigned one or more semantic tags reflecting its reasoning domain, visual theme, and task type. Statistical analysis of the tag distribution indicates that the curated corpus covers a broad range of domains and cognitive skills, including but not limited to:
\begin{itemize}
    \item \textbf{Visual Understanding:} This includes object recognition (\texttt{object}), attribute parsing (\texttt{attribute}), and holistic scene comprehension (\texttt{scene}).
    \item \textbf{Spatial Reasoning:} Tags represent positional and relational understanding such as spatial relations (\texttt{spatial}, \texttt{position}, \texttt{layout}), directionality (\texttt{direction}), and coordinate-based inference (\texttt{coordinate}).
    \item \textbf{Chart and Table Interpretation:} We support multiple visualization formats, reflected in tags such as \texttt{chart}, \texttt{graph}, \texttt{table}, \texttt{bar chart}, \texttt{pie chart}, and \texttt{scatter plot}, as well as related meta-information (\texttt{axis}, \texttt{legend}, \texttt{label}, \texttt{caption}).
    \item \textbf{Logical and Mathematical Reasoning:} The dataset includes tags for counting (\texttt{count}), comparison (\texttt{comparison}), and broader mathematical and statistical reasoning (\texttt{math}, \texttt{algebra}, \texttt{geometry}, \texttt{percentage}, \texttt{ratio}, \texttt{average}).
    \item \textbf{Domain-Specific Knowledge:} Coverage extends to science (\texttt{physics}, \texttt{chemistry}, \texttt{biology}), technology (\texttt{code}, \texttt{algorithm}, \texttt{electronics}), and humanities (\texttt{history}, \texttt{politics}, \texttt{literature}).
    \item \textbf{Reflective and Error-Aware Reasoning:} Tags emphasize verification (\texttt{verification}, \texttt{fact-check}), error identification (\texttt{error}, \texttt{discrepancy}), and attention to detail (\texttt{precision}, \texttt{detail}).
\end{itemize}

These semantic tags support systematic sampling, targeted curriculum design, and analysis across diverse reasoning tasks and domains.

\section{Case Study}
\label{case_study}

We qualitatively compare source annotations from FineVision with the structured annotations produced by \textsc{OmniThoughtVis}. These examples are intended to illustrate how structured reasoning traces can make supervision more auditable, explicit, and useful for downstream distillation. In many cases, the source annotations provide concise conclusions or coarse descriptions, while our pipeline decomposes the task into explicit reasoning steps (e.g., element identification, subgoal reasoning, evidence checking, and exception identification). This structure can improve interpretability and make it easier to inspect or reuse the resulting supervision.

\textbf{Case 1 (Scientific concept grounding).}
The FineVision annotation gives a brief justification for ``ionic'' bonding, but remains relatively coarse. Our annotation follows a structured checklist (entity identification $\rightarrow$ bonding-type analysis $\rightarrow$ dominance judgment $\rightarrow$ structural verification), explicitly connecting the final answer to observable cues such as metal--oxygen composition, charged oxygen, and coordination pattern.

\textbf{Case 2 (Geometry problem; error correction).}
The FineVision annotation outputs an option letter (A) without intermediate verification. Our structured annotation decomposes the solution into geometry subgoals (triangle properties $\rightarrow$ perpendicular-from-center theorem $\rightarrow$ angle/arc relationships), providing a traceable derivation that also recovers the correct answer.

\textbf{Case 3 (Meme understanding with multi-dimensional affect).}
The FineVision annotation summarizes the meme with coarse labels (e.g., ``funny'' and ``positive''). Our annotation instead allocates one reasoning step to each requested dimension (humor, sarcasm, offensiveness, motivational quality, and sentiment), tying the judgment more explicitly to textual and visual evidence.

\textbf{Case 4 (Chart reasoning; error correction).}
The FineVision annotation answers ``Yes'' without supporting comparison. Our annotation performs a more explicit curve-by-curve analysis followed by an overall comparison, making the visual evidence behind the answer easier to verify.

\textbf{Case 5 (Fine-grained visual search in a cluttered scene).}
The FineVision annotation points to a single candidate using a brief description. Our annotation adopts an \textbf{enumerate-then-identify} strategy: it first lists visible individuals and their helmet status, and only then identifies the unique exception. This yields more explicit evidence coverage for ``find-the-exception'' tasks.

\begin{figure*}[t]
\begin{casestudy}[Case 1]
\begin{center}
    \fcolorbox{caseheader!20}{questionbg!30}{%
        \includegraphics[height=6cm]{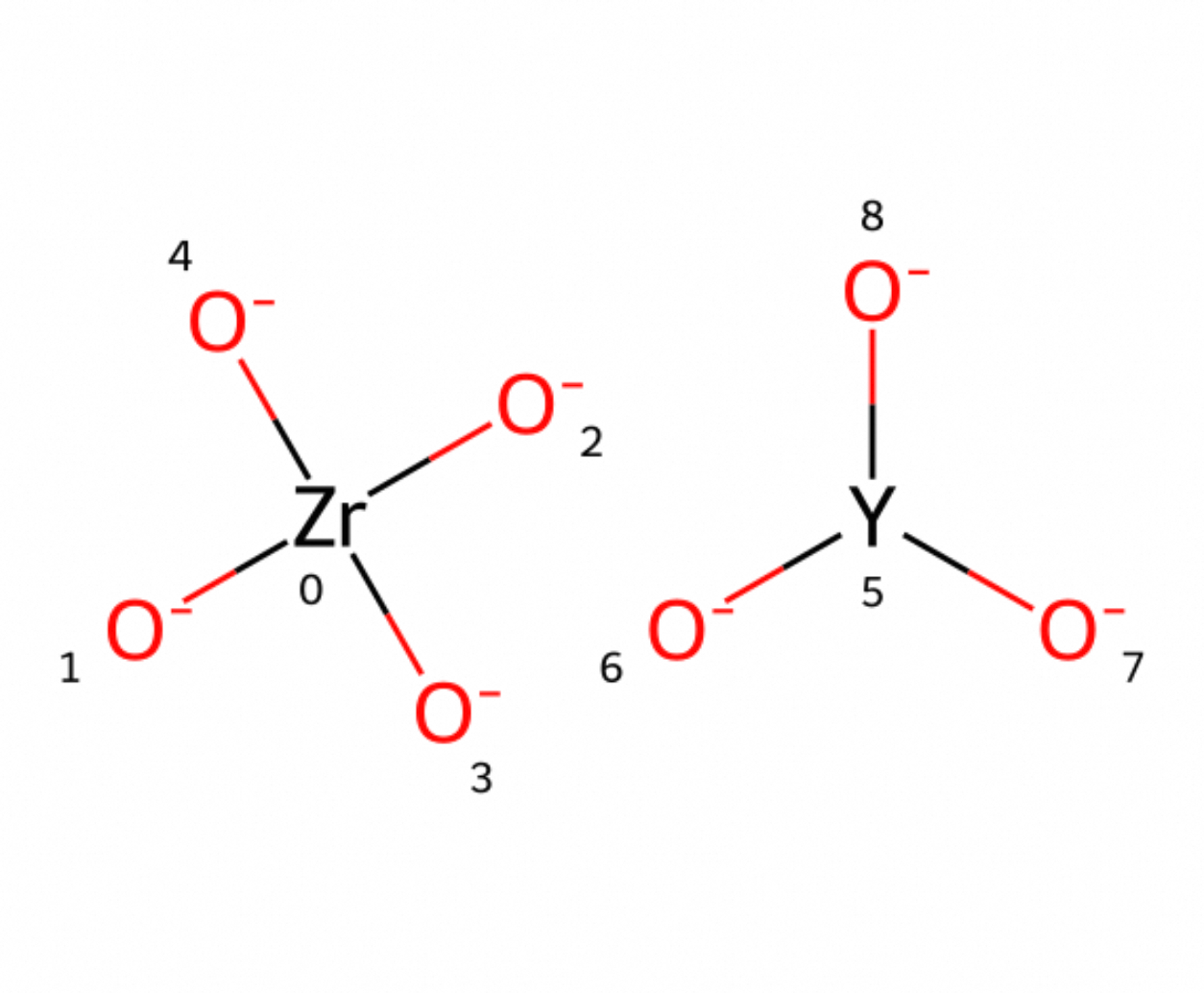}%
    }
\end{center}

\vspace{4pt}

\begin{questionbox}
\textbf{Question:} Does this compound exhibit ionic or covalent bonding?
\end{questionbox}

\vspace{4pt}

\begin{originalbox}
\textbf{\textcolor{originalframe}{$\blacksquare$~Original Annotation (FineVision):}}
The interactions between zirconium and oxygen in this structure indicate ionic bonding characteristics, particularly due to the presence of charged oxygen. \textbf{Answer:} ionic
\end{originalbox}

\vspace{4pt}

\begin{oursbox}
\textbf{\textcolor{oursframe}{$\bigstar$~Structured Annotation (Ours):}}\\[4pt]
\thinktag\\[2pt]
To determine whether the compound exhibits ionic or covalent bonding, we need to analyze the nature of the bonds between the atoms in the structure.

\vspace{3pt}
\textcolor{stepcolor}{\textbf{Step 1: Identify the elements involved.}} The structure contains zirconium (Zr), yttrium (Y), and oxygen (O). Oxygen is highly electronegative, while Zr and Y are metals.

\vspace{2pt}
\textcolor{stepcolor}{\textbf{Step 2: Analyze metal--oxygen bonding.}} Zr and Y are transition metals bonded to oxygen atoms. Oxygen typically forms anions (O$^-$) in such compounds. The presence of O$^-$ ions suggests electrons are transferred from the metals to oxygen, forming ionic bonds.

\vspace{2pt}
\textcolor{stepcolor}{\textbf{Step 3: Evaluate the dominant bonding type.}} In oxides of metals like Zr and Y, the bonding is primarily ionic due to the large difference in electronegativity. Although some covalent character may exist due to polarization, the dominant bonding type remains ionic.

\vspace{2pt}
\textcolor{stepcolor}{\textbf{Step 4: Structural verification.}} The structure shows Zr and Y each bonded to four O$^-$ ions, typical for ionic oxides. The arrangement resembles a perovskite-like structure with metal cations surrounded by oxygen anions, consistent with ionic bonding.

\vspace{2pt}
\textcolor{stepcolor}{\textbf{Conclusion:}} Given the involvement of metals (Zr, Y) and oxygen, and the presence of O$^-$ ions, the bonding is predominantly ionic.\\[2pt]
\thinktagclose

\vspace{4pt}
\answertag

~Ionic bonding~

\answertagclose
\end{oursbox}

\end{casestudy}
\end{figure*}


\begin{figure*}[t]
\begin{casestudy}[Case 2]

\begin{center}
    \fcolorbox{caseheader!20}{questionbg!30}{%
        \includegraphics[height=6cm]{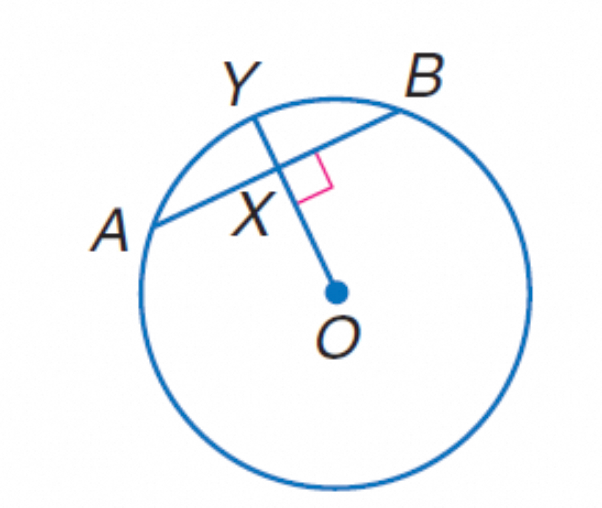}%
    }
\end{center}

\vspace{4pt}

\begin{questionbox}
\textbf{Question:} Answer the mathematical geometry problem and directly provide the correct option letter.\\[2pt]
Circle $O$ has a radius of 10, $AB = 10$ and $m\,\overarc{AB} = 60$. Find $m\,\overarc{AX}$.\\[2pt]
\textbf{Choices:} A: 5 \quad B: 10 \quad C: 20 \quad D: 30
\end{questionbox}

\vspace{4pt}

\begin{originalbox}
\textbf{\textcolor{originalframe}{$\blacksquare$~Original Annotation (FineVision):}} A
\end{originalbox}

\vspace{4pt}

\begin{oursbox}
\textbf{\textcolor{oursframe}{$\bigstar$~Structured Annotation (Ours):}}\\[4pt]
\thinktag\\[2pt]
We are given: Circle $O$ has a radius of 10, $AB = 10$, and $m\,\overarc{AB} = 60^\circ$. We need to find $m\,\overarc{AX}$.

\vspace{3pt}
\textcolor{stepcolor}{\textbf{Step 1: Analyze triangle $OAB$.}} Since $AB$ is a chord and $OA = OB = 10$ (radii), and $AB = 10$, triangle $OAB$ has all sides equal to 10~--- it is an equilateral triangle. Therefore $\angle AOB = 60^\circ$, consistent with $m\,\overarc{AB} = 60^\circ$.

\vspace{2pt}
\textcolor{stepcolor}{\textbf{Step 2: Identify point $X$.}} Point $X$ is the foot of the perpendicular from center $O$ to chord $AB$. By the perpendicular-from-center theorem, $OX \perp AB$ bisects the chord, so $AX = XB = AB/2 = 5$.

\vspace{2pt}
\textcolor{stepcolor}{\textbf{Step 3: Compute the arc.}} Since $OX$ bisects both chord $AB$ and central angle $\angle AOB$, we have $\angle AOX = \angle BOX = 60^\circ / 2 = 30^\circ$. Arc $AX$ corresponds to central angle $\angle AOX$, so $m\,\overarc{AX} = 30^\circ$.

\vspace{2pt}
\textcolor{stepcolor}{\textbf{Conclusion:}} $m\,\overarc{AX} = 30$. The correct answer is \textbf{D}.\\[2pt]
\thinktagclose

\vspace{4pt}
\answertag

D

\answertagclose
\end{oursbox}
\end{casestudy}
\end{figure*}


\begin{figure*}[t]
\begin{casestudy}[Case 3]

\begin{center}
    \fcolorbox{caseheader!20}{questionbg!30}{%
        \includegraphics[height=4cm]{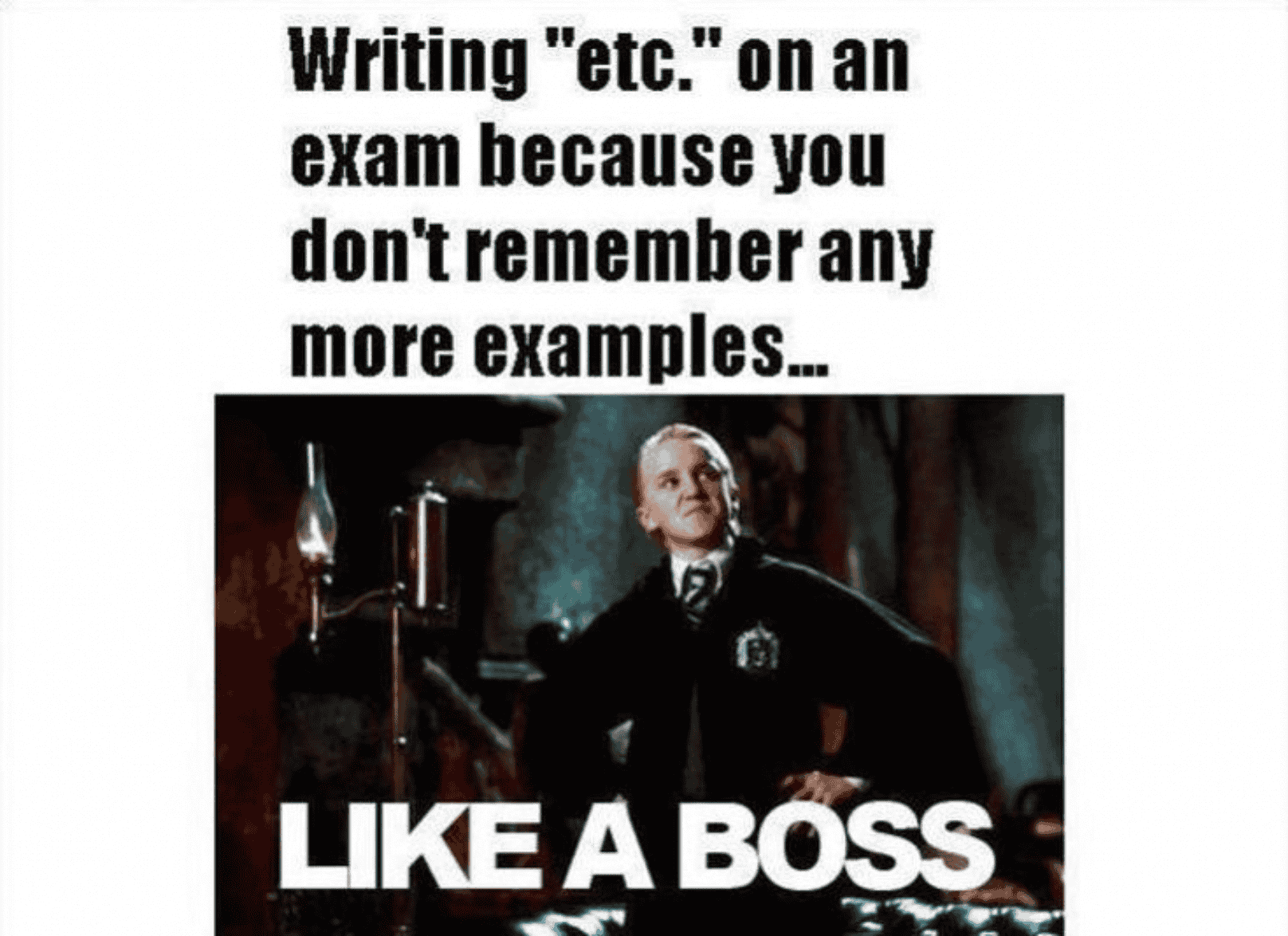}%
    }
\end{center}

\vspace{4pt}

\begin{questionbox}
\textbf{Question:} Break down this meme's emotional profile: humor level, sarcasm, offensiveness, motivational qualities, and sentiment.
\end{questionbox}

\vspace{4pt}

\begin{originalbox}
\textbf{\textcolor{originalframe}{$\blacksquare$~Original Annotation (FineVision):}}
This meme is funny. It is generally sarcastic, not offensive, and not motivational. The overall sentiment is positive.
\end{originalbox}

\vspace{4pt}

\begin{oursbox}
\textbf{\textcolor{oursframe}{$\bigstar$~Structured Annotation (Ours):}}\\[4pt]
\thinktag\\[2pt]
To analyze the emotional profile of this meme, let's break it down step by step:

\vspace{3pt}
\textcolor{stepcolor}{\textbf{1. Humor Level.}} The humor stems from relatability and irony. Many students have experienced the frustration of being unable to recall examples during an exam and resorting to writing ``etc.'' as a placeholder. The image of Draco Malfoy, known for his smug demeanor, exaggerates this behavior. This contrast between a common academic struggle and a character known for superiority creates a light-hearted, situational humor.

\vspace{2pt}
\textcolor{stepcolor}{\textbf{2. Sarcasm.}} There is a strong element of sarcasm. The phrase ``LIKE A BOSS'' is typically used to mock someone acting overly confident. Here, it is applied to a student who pretends to be knowledgeable by using ``etc.,'' highlighting the absurdity of feigning expertise.

\vspace{2pt}
\textcolor{stepcolor}{\textbf{3. Offensiveness.}} The meme is not offensive. It does not target any specific group, individual, or sensitive topic. Instead, it pokes fun at a universal student experience in a non-malicious way.

\vspace{2pt}
\textcolor{stepcolor}{\textbf{4. Motivational Qualities.}} The meme lacks motivational qualities. It mocks a common academic shortcut rather than inspiring positive behavior, making the audience laugh at their own past struggles.

\vspace{2pt}
\textcolor{stepcolor}{\textbf{5. Sentiment.}} The overall sentiment is negative but in a playful, ironic way. It expresses frustration with academic limitations through humor and exaggeration. The tone is self-deprecating and slightly mocking, but not mean-spirited.

\vspace{2pt}
\textcolor{stepcolor}{\textbf{Summary:}} The meme uses relatable academic anxiety, exaggerated confidence, and sarcasm to create humor without crossing into offensiveness.\\[2pt]
\thinktagclose

\vspace{4pt}
\answertag\\[1pt]
Humor Level: High (relatable, ironic, and exaggerated); Sarcasm: High (mocking the pretense of knowledge); Offensiveness: Low (no targeted insults or sensitive topics); Motivational Qualities: None (mocks rather than inspires); Sentiment: Playful negativity (self-deprecating humor with irony)\\[1pt]
\answertagclose
\end{oursbox}

\end{casestudy}
\end{figure*}


\begin{figure*}[t]
\begin{casestudy}[Case 4]

\begin{center}
    \fcolorbox{caseheader!20}{questionbg!30}{%
        \includegraphics[height=6cm]{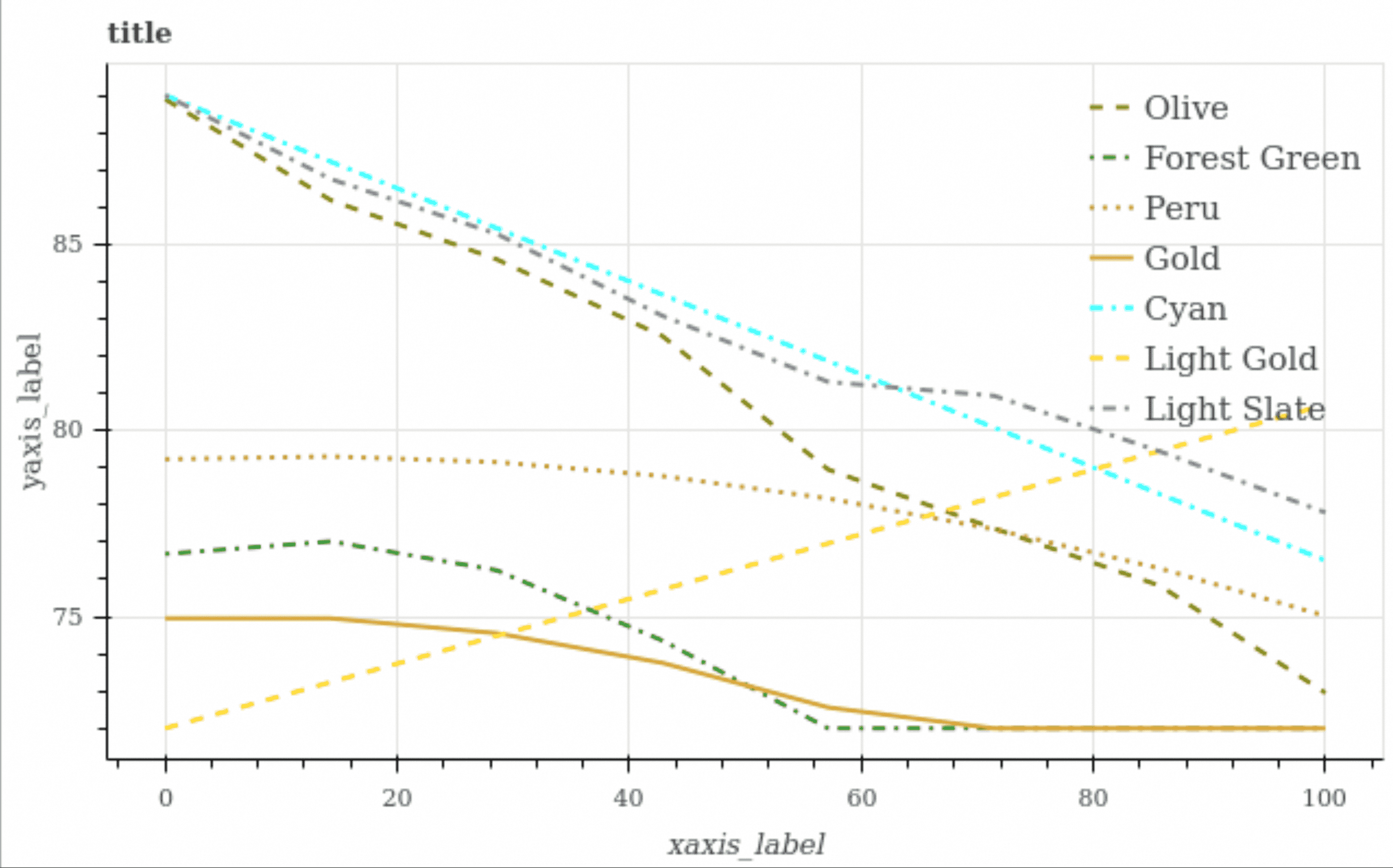}%
    }
\end{center}

\vspace{4pt}

\begin{questionbox}
\textbf{Question:} Does Forest Green have the maximum area under the curve? Answer yes or no.
\end{questionbox}

\vspace{4pt}

\begin{originalbox}
\textbf{\textcolor{originalframe}{$\blacksquare$~Original Annotation (FineVision):}} Yes.
\end{originalbox}

\vspace{4pt}

\begin{oursbox}
\textbf{\textcolor{oursframe}{$\bigstar$~Structured Annotation (Ours):}}\\[4pt]
\thinktag\\[2pt]
To determine whether Forest Green has the maximum area under the curve, we need to compare the areas under all lines in the graph. The area under a curve is proportional to how high the line stays above the $x$-axis over the entire range.

\vspace{3pt}
\textcolor{stepcolor}{\textbf{Step 1: Analyze each line.}}
\textbf{Cyan} starts at the highest value ($\sim$98) and decreases steadily but remains relatively high throughout.
\textbf{Light Slate} also starts high and decreases gradually, staying above most other lines except Cyan.
\textbf{Olive} starts high but drops sharply after the first few points.
\textbf{Peru} remains fairly flat around 78--79.
\textbf{Gold} starts low and increases slowly.
\textbf{Light Gold} starts low and increases steadily, crossing some lines but not reaching the top.
\textbf{Forest Green} starts around 75, peaks slightly, then drops sharply and ends near zero.

\vspace{2pt}
\textcolor{stepcolor}{\textbf{Step 2: Compare areas.}} Cyan and Light Slate are the two highest lines throughout the range. Cyan starts higher and maintains a higher position for most of the range. Forest Green, while starting moderately high, drops quickly and ends near zero, so its area is significantly smaller.

\vspace{2pt}
\textcolor{stepcolor}{\textbf{Conclusion:}} Forest Green does \textbf{not} have the maximum area under the curve. Cyan has the largest area due to its consistently high values across the entire range.\\[2pt]
\thinktagclose

\vspace{4pt}
\answertag

No

\answertagclose
\end{oursbox}

\end{casestudy}
\end{figure*}


\begin{figure*}[t]
\begin{casestudy}[Case 5]

\begin{center}
    \fcolorbox{caseheader!20}{questionbg!30}{%
        \includegraphics[height=7cm]{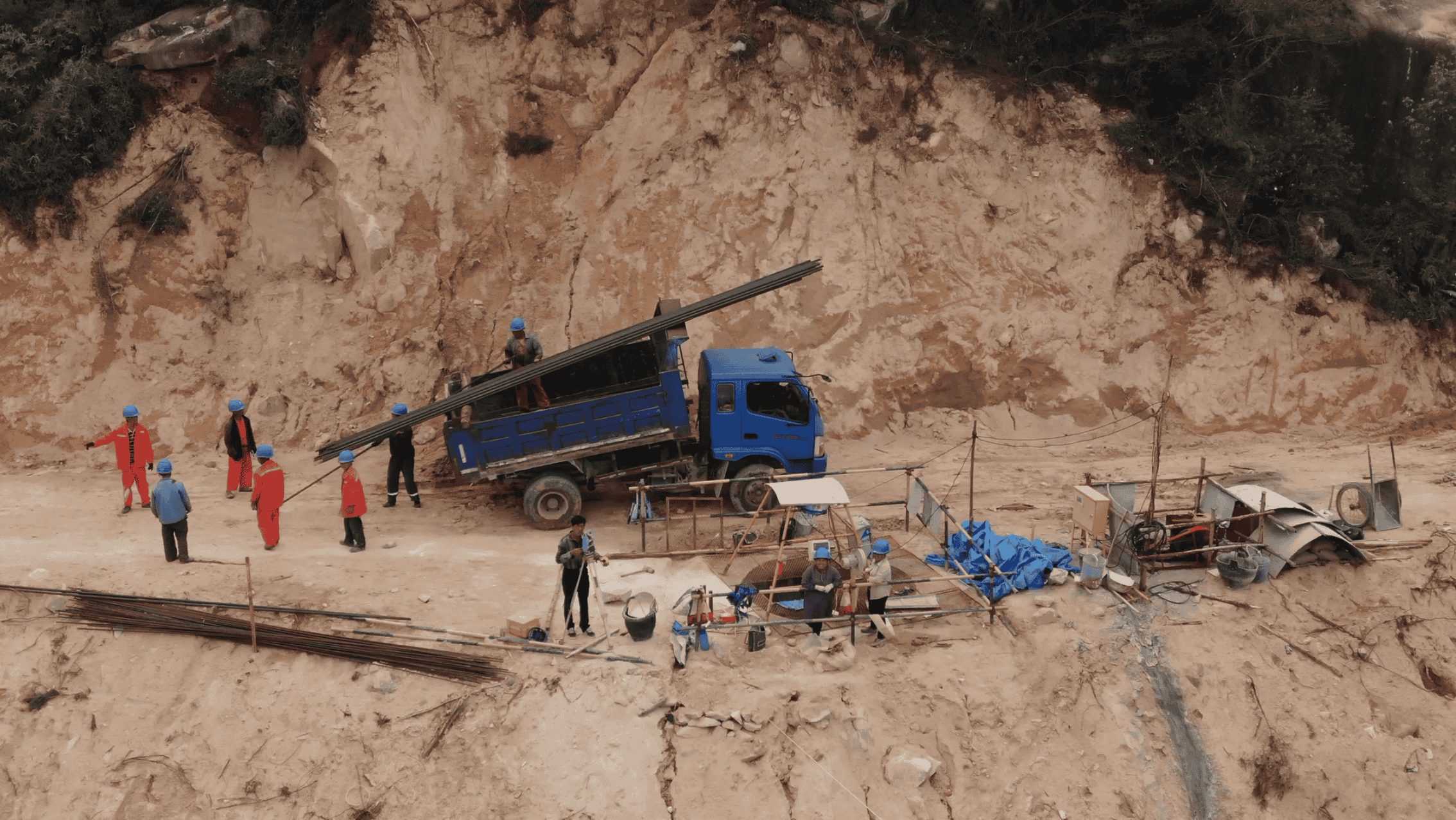}%
    }
\end{center}

\vspace{4pt}

\begin{questionbox}
\textbf{Question:} Point out the person in the picture who is not wearing a helmet.
\end{questionbox}

\vspace{4pt}

\begin{originalbox}
\textbf{\textcolor{originalframe}{$\blacksquare$~Original Annotation (FineVision):}}
In the image, there is a person wearing a pink shirt and dark pants, standing near the center of the image to the right of a blue truck, who is not wearing a helmet. This individual is in contrast to the others around who are wearing helmets, presumably for safety reasons at this construction or work site.
\end{originalbox}

\vspace{4pt}

\begin{oursbox}
\textbf{\textcolor{oursframe}{$\bigstar$~Structured Annotation (Ours):}}\\[4pt]
\thinktag\\[2pt]
To determine who is not wearing a helmet, I will examine each person in the image:

\vspace{3pt}
\textcolor{stepcolor}{\textbf{Step 1: Enumerate all individuals.}}

(1)~Person on the left side of the blue truck: orange outfit, blue helmet.

(2)~Person next to them: orange outfit, blue helmet.

(3)~Person in the center-left: red outfit, blue helmet.

(4)~Person in the center-right: dark outfit, blue helmet.

(5)~Person near the bottom center: gray shirt, black pants, \textbf{no helmet}; holding a measuring tape.

(6)~Person near the right side under the makeshift shelter: blue helmet.

(7)~Person near the far right, partially obscured: blue helmet.

\vspace{2pt}
\textcolor{stepcolor}{\textbf{Step 2: Identify the exception.}} 

After reviewing all individuals, only person~(5) in the gray shirt is not wearing a helmet.\\[2pt]
\thinktagclose

\vspace{4pt}
\answertag

The person in the gray shirt near the bottom center of the image is not wearing a helmet.

\answertagclose
\end{oursbox}

\end{casestudy}
\end{figure*}


\begin{thebibliography}{34}
\expandafter\ifx\csname natexlab\endcsname\relax\def\natexlab#1{#1}\fi

\bibitem[{Bai et~al.(2025{\natexlab{a}})Bai, Cai, Chen, Chen, Chen, Cheng, Deng, Ding, Gao, Ge, Ge, Guo, Huang, Huang, Huang, Hui, Jiang, Li, Li, Li, Li, Lin, Lin, Liu, Liu, Liu, Liu, Liu, Liu, Lu, Luo, Lv, Men, Meng, Ren, Ren, Song, Sun, Tang, Tu, Wan, Wang, Wang, Wang, Wang, Xie, Xu, Xu, Xu, Yang, Yang, Yang, Yang, Yu, Zhang, Zhang, Zhang, Zheng, Zhong, Zhou, Zhou, Zhou, Zhu, and Zhu}]{qwen3vl}
Shuai Bai, Yuxuan Cai, Ruizhe Chen, Keqin Chen, Xionghui Chen, Zesen Cheng, Lianghao Deng, Wei Ding, Chang Gao, Chunjiang Ge, Wenbin Ge, Zhifang Guo, Qidong Huang, Jie Huang, Fei Huang, Binyuan Hui, Shutong Jiang, Zhaohai Li, Mingsheng Li, Mei Li, Kaixin Li, Zicheng Lin, Junyang Lin, Xuejing Liu, Jiawei Liu, Chenglong Liu, Yang Liu, Dayiheng Liu, Shixuan Liu, Dunjie Lu, Ruilin Luo, Chenxu Lv, Rui Men, Lingchen Meng, Xuancheng Ren, Xingzhang Ren, Sibo Song, Yuchong Sun, Jun Tang, Jianhong Tu, Jianqiang Wan, Peng Wang, Pengfei Wang, Qiuyue Wang, Yuxuan Wang, Tianbao Xie, Yiheng Xu, Haiyang Xu, Jin Xu, Zhibo Yang, Mingkun Yang, Jianxin Yang, An~Yang, Bowen Yu, Fei Zhang, Hang Zhang, Xi~Zhang, Bo~Zheng, Humen Zhong, Jingren Zhou, Fan Zhou, Jing Zhou, Yuanzhi Zhu, and Ke~Zhu. 2025{\natexlab{a}}.
\newblock \href {http://arxiv.org/abs/2511.21631} {Qwen3-vl technical report}.

\bibitem[{Bai et~al.(2025{\natexlab{b}})Bai, Chen, Liu, Wang, Ge, Song, Dang, Wang, Wang, Tang, Zhong, Zhu, Yang, Li, Wan, Wang, Ding, Fu, Xu, Ye, Zhang, Xie, Cheng, Zhang, Yang, Xu, and Lin}]{bai2025qwen25vltechnicalreport}
Shuai Bai, Keqin Chen, Xuejing Liu, Jialin Wang, Wenbin Ge, Sibo Song, Kai Dang, Peng Wang, Shijie Wang, Jun Tang, Humen Zhong, Yuanzhi Zhu, Mingkun Yang, Zhaohai Li, Jianqiang Wan, Pengfei Wang, Wei Ding, Zheren Fu, Yiheng Xu, Jiabo Ye, Xi~Zhang, Tianbao Xie, Zesen Cheng, Hang Zhang, Zhibo Yang, Haiyang Xu, and Junyang Lin. 2025{\natexlab{b}}.
\newblock \href {http://arxiv.org/abs/2502.13923} {Qwen2.5-vl technical report}.

\bibitem[{Cai et~al.(2025)Cai, Wang, Yan, Huang, and Fang}]{cai2025reasoning}
Wenrui Cai, Chengyu Wang, Junbing Yan, Jun Huang, and Xiangzhong Fang. 2025.
\newblock Reasoning with omnithought: A large cot dataset with verbosity and cognitive difficulty annotations.
\newblock \emph{arXiv preprint arXiv:2505.10937}.

\bibitem[{Chen et~al.(2024{\natexlab{a}})Chen, Li, Dong, Zhang, He, Wang, Zhao, and Lin}]{ShareGPT4V}
Lin Chen, Jinsong Li, Xiaoyi Dong, Pan Zhang, Conghui He, Jiaqi Wang, Feng Zhao, and Dahua Lin. 2024{\natexlab{a}}.
\newblock \href {https://doi.org/10.1007/978-3-031-72643-9_22} {Sharegpt4v: Improving large multi-modal models with better captions}.
\newblock In \emph{Computer Vision -- ECCV 2024: 18th European Conference, Milan, Italy, September 29--October 4, 2024, Proceedings, Part XVII}, pages 370--387, Berlin, Heidelberg. Springer-Verlag.

\bibitem[{Chen et~al.(2024{\natexlab{b}})Chen, Li, Dong, Zhang, Zang, Chen, Duan, Wang, Qiao, Lin, and Zhao}]{mmstar}
Lin Chen, Jinsong Li, Xiaoyi Dong, Pan Zhang, Yuhang Zang, Zehui Chen, Haodong Duan, Jiaqi Wang, Yu~Qiao, Dahua Lin, and Feng Zhao. 2024{\natexlab{b}}.
\newblock \href {https://openreview.net/forum?id=evP9mxNNxJ} {Are we on the right way for evaluating large vision-language models?}
\newblock In \emph{The Thirty-eighth Annual Conference on Neural Information Processing Systems}.

\bibitem[{Chung et~al.(2024)Chung, Hou, Longpre, Zoph, Tay, Fedus, Li, Wang, Dehghani, Brahma et~al.}]{10.5555/3722577.3722647}
Hyung~Won Chung, Le~Hou, Shayne Longpre, Barret Zoph, Yi~Tay, William Fedus, Yunxuan Li, Xuezhi Wang, Mostafa Dehghani, Siddhartha Brahma, et~al. 2024.
\newblock Scaling instruction-finetuned language models.
\newblock \emph{Journal of Machine Learning Research}, 25(70):1--53.

\bibitem[{DeepSeek-AI(2025)}]{DeepSeek-R1}
DeepSeek-AI. 2025.
\newblock \href {https://doi.org/10.1038/s41586-025-09422-z} {Deepseek-r1 incentivizes reasoning in llms through reinforcement learning}.
\newblock \emph{Nature}, 645(8081):633--638.

\bibitem[{Ester et~al.(1996)Ester, Kriegel, Sander, and Xu}]{DBSCAN}
Martin Ester, Hans-Peter Kriegel, J\"{o}rg Sander, and Xiaowei Xu. 1996.
\newblock A density-based algorithm for discovering clusters in large spatial databases with noise.
\newblock In \emph{Proceedings of the Second International Conference on Knowledge Discovery and Data Mining}, KDD'96, pages 226--231. AAAI Press.

\bibitem[{Guha et~al.(2026)Guha, Marten, Keh, Raoof, Smyrnis, Bansal, Nezhurina, Mercat, Vu, Sprague, Suvarna, Feuer, Chen, Khan, Frankel, Grover, Choi, Muennighoff, Su, Zhao, Yang, Pimpalgaonkar, sharma, Ji, Deng, Pratt, Ramanujan, Saad-Falcon, Acharya, Li, Dave, Albalak, Arora, Wulfe, Hegde, Durrett, Oh, Bansal, Gabriel, Grover, Chang, Shankar, Gokaslan, Merrill, Hashimoto, Choi, Jitsev, Heckel, Sathiamoorthy, Dimakis, and Schmidt}]{openthoughts}
Etash~Kumar Guha, Ryan Marten, Sedrick Keh, Negin Raoof, Georgios Smyrnis, Hritik Bansal, Marianna Nezhurina, Jean Mercat, Trung Vu, Zayne~Rea Sprague, Ashima Suvarna, Benjamin Feuer, Leon~Liangyu Chen, Zaid Khan, Eric Frankel, Sachin Grover, Caroline Choi, Niklas Muennighoff, Shiye Su, Wanjia Zhao, John Yang, Shreyas Pimpalgaonkar, Kartik sharma, Charlie Cheng-Jie Ji, Yichuan Deng, Sarah~M Pratt, Vivek Ramanujan, Jon Saad-Falcon, Stutee Acharya, Jeffrey Li, Achal Dave, Alon Albalak, Kushal Arora, Blake Wulfe, Chinmay Hegde, Greg Durrett, Sewoong Oh, Mohit Bansal, Saadia Gabriel, Aditya Grover, Kai-Wei Chang, Vaishaal Shankar, Aaron Gokaslan, Mike~A Merrill, Tatsunori Hashimoto, Yejin Choi, Jenia Jitsev, Reinhard Heckel, Maheswaran Sathiamoorthy, Alex Dimakis, and Ludwig Schmidt. 2026.
\newblock \href {https://openreview.net/forum?id=7xjoTuaNmN} {Openthoughts: Data recipes for reasoning models}.
\newblock In \emph{The Fourteenth International Conference on Learning Representations}.

\bibitem[{Hinton et~al.(2015)Hinton, Vinyals, and Dean}]{hinton2015distillingknowledgeneuralnetwork}
Geoffrey Hinton, Oriol Vinyals, and Jeff Dean. 2015.
\newblock Distilling the knowledge in a neural network.
\newblock \emph{arXiv preprint arXiv:1503.02531}.

\bibitem[{Ho et~al.(2023)Ho, Schmid, and Yun}]{ho-etal-2023-large}
Namgyu Ho, Laura Schmid, and Se-Young Yun. 2023.
\newblock \href {https://doi.org/10.18653/v1/2023.acl-long.830} {Large language models are reasoning teachers}.
\newblock In \emph{Proceedings of the 61st Annual Meeting of the Association for Computational Linguistics (Volume 1: Long Papers)}, pages 14852--14882, Toronto, Canada. Association for Computational Linguistics.

\bibitem[{Hsieh et~al.(2023)Hsieh, Li, Yeh, Nakhost, Fujii, Ratner, Krishna, Lee, and Pfister}]{hsieh-etal-2023-distilling}
Cheng-Yu Hsieh, Chun-Liang Li, Chih-kuan Yeh, Hootan Nakhost, Yasuhisa Fujii, Alex Ratner, Ranjay Krishna, Chen-Yu Lee, and Tomas Pfister. 2023.
\newblock \href {https://doi.org/10.18653/v1/2023.findings-acl.507} {Distilling step-by-step! outperforming larger language models with less training data and smaller model sizes}.
\newblock In \emph{Findings of the Association for Computational Linguistics: ACL 2023}, pages 8003--8017, Toronto, Canada. Association for Computational Linguistics.

\bibitem[{Kembhavi et~al.(2016)Kembhavi, Salvato, Kolve, Seo, Hajishirzi, and Farhadi}]{ai2d}
Aniruddha Kembhavi, Mike Salvato, Eric Kolve, Minjoon Seo, Hannaneh Hajishirzi, and Ali Farhadi. 2016.
\newblock A diagram is worth a dozen images.
\newblock In \emph{European conference on computer vision}, pages 235--251. Springer.

\bibitem[{Li et~al.(2025)Li, Zhang, Guo, Zhang, Li, Zhang, Zhang, Zhang, Li, Liu, and Li}]{li2025llavaonevision}
Bo~Li, Yuanhan Zhang, Dong Guo, Renrui Zhang, Feng Li, Hao Zhang, Kaichen Zhang, Peiyuan Zhang, Yanwei Li, Ziwei Liu, and Chunyuan Li. 2025.
\newblock \href {https://openreview.net/forum?id=zKv8qULV6n} {{LL}a{VA}-onevision: Easy visual task transfer}.
\newblock \emph{Transactions on Machine Learning Research}.

\bibitem[{Liu et~al.(2023)Liu, Li, Wu, and Lee}]{liu2023llava}
Haotian Liu, Chunyuan Li, Qingyang Wu, and Yong~Jae Lee. 2023.
\newblock Visual instruction tuning.
\newblock \emph{Advances in neural information processing systems}, 36:34892--34916.

\bibitem[{Liu et~al.(2024)Liu, Duan, Zhang, Li, Zhang, Zhao, Yuan, Wang, He, Liu, Chen, and Lin}]{mmbench}
Yuan Liu, Haodong Duan, Yuanhan Zhang, Bo~Li, Songyang Zhang, Wangbo Zhao, Yike Yuan, Jiaqi Wang, Conghui He, Ziwei Liu, Kai Chen, and Dahua Lin. 2024.
\newblock \href {https://doi.org/10.1007/978-3-031-72658-3_13} {Mmbench: Is your multi-modal model an all-around player?}
\newblock In \emph{Computer Vision -- ECCV 2024: 18th European Conference, Milan, Italy, September 29--October 4, 2024, Proceedings, Part VI}, pages 216--233, Berlin, Heidelberg. Springer-Verlag.

\bibitem[{Lu et~al.(2024)Lu, Bansal, Xia, Liu, Li, Hajishirzi, Cheng, Chang, Galley, and Gao}]{mathvista}
Pan Lu, Hritik Bansal, Tony Xia, Jiacheng Liu, Chunyuan Li, Hannaneh Hajishirzi, Hao Cheng, Kai-Wei Chang, Michel Galley, and Jianfeng Gao. 2024.
\newblock \href {https://openreview.net/forum?id=KUNzEQMWU7} {Mathvista: Evaluating mathematical reasoning of foundation models in visual contexts}.
\newblock In \emph{The Twelfth International Conference on Learning Representations}.

\bibitem[{Lu et~al.(2022)Lu, Mishra, Xia, Qiu, Chang, Zhu, Tafjord, Clark, and Kalyan}]{scienceQA}
Pan Lu, Swaroop Mishra, Tanglin Xia, Liang Qiu, Kai-Wei Chang, Song-Chun Zhu, Oyvind Tafjord, Peter Clark, and Ashwin Kalyan. 2022.
\newblock \href {https://proceedings.neurips.cc/paper_files/paper/2022/file/11332b6b6cf4485b84afadb1352d3a9a-Paper-Conference.pdf} {Learn to explain: Multimodal reasoning via thought chains for science question answering}.
\newblock In \emph{Advances in Neural Information Processing Systems}, volume~35, pages 2507--2521. Curran Associates, Inc.

\bibitem[{Mukherjee et~al.(2023)Mukherjee, Mitra, Jawahar, Agarwal, Palangi, and Awadallah}]{mukherjee2023orca}
Subhabrata Mukherjee, Arindam Mitra, Ganesh Jawahar, Sahaj Agarwal, Hamid Palangi, and Ahmed Awadallah. 2023.
\newblock Orca: Progressive learning from complex explanation traces of gpt-4.
\newblock \emph{arXiv preprint arXiv:2306.02707}.

\bibitem[{Tong et~al.(2024)Tong, II, Wu, Woo, IYER, Akula, Yang, Yang, Middepogu, Wang, Pan, Fergus, LeCun, and Xie}]{tong2024cambrian}
Shengbang Tong, Ellis L~Brown II, Penghao Wu, Sanghyun Woo, ADITHYA~JAIRAM IYER, Sai~Charitha Akula, Shusheng Yang, Jihan Yang, Manoj Middepogu, Ziteng Wang, Xichen Pan, Rob Fergus, Yann LeCun, and Saining Xie. 2024.
\newblock \href {https://openreview.net/forum?id=Vi8AepAXGy} {Cambrian-1: A fully open, vision-centric exploration of multimodal {LLM}s}.
\newblock In \emph{The Thirty-eighth Annual Conference on Neural Information Processing Systems}.

\bibitem[{Toshniwal et~al.(2024)Toshniwal, Moshkov, Narenthiran, Gitman, Jia, and Gitman}]{toshniwal2024openmathinstruct118millionmath}
Shubham Toshniwal, Ivan Moshkov, Sean Narenthiran, Daria Gitman, Fei Jia, and Igor Gitman. 2024.
\newblock Openmathinstruct-1: A 1.8 million math instruction tuning dataset.
\newblock \emph{Advances in Neural Information Processing Systems}, 37:34737--34774.

\bibitem[{Wang et~al.(2025)Wang, Yan, Yue, and Huang}]{wang2025distilqwen2}
Chengyu Wang, Junbing Yan, Yuanhao Yue, and Jun Huang. 2025.
\newblock Distilqwen2.5: Industrial practices of training distilled open lightweight language models.
\newblock \emph{arXiv preprint arXiv:2504.15027}.

\bibitem[{Wang et~al.(2024)Wang, Pan, Shi, Lu, Ren, Zhou, Zhan, and Li}]{mathvision}
Ke~Wang, Junting Pan, Weikang Shi, Zimu Lu, Houxing Ren, Aojun Zhou, Mingjie Zhan, and Hongsheng Li. 2024.
\newblock Measuring multimodal mathematical reasoning with math-vision dataset.
\newblock \emph{Advances in Neural Information Processing Systems}, 37:95095--95169.

\bibitem[{Wei et~al.(2022)Wei, Wang, Schuurmans, Bosma, Xia, Chi, Le, Zhou et~al.}]{cot}
Jason Wei, Xuezhi Wang, Dale Schuurmans, Maarten Bosma, Fei Xia, Ed~Chi, Quoc~V Le, Denny Zhou, et~al. 2022.
\newblock Chain-of-thought prompting elicits reasoning in large language models.
\newblock \emph{Advances in neural information processing systems}, 35:24824--24837.

\bibitem[{Wiedmann et~al.(2025)Wiedmann, Zohar, Mahla, Wang, Li, Frere, von Werra, Gosthipaty, and Marafioti}]{finevision}
Luis Wiedmann, Orr Zohar, Amir Mahla, Xiaohan Wang, Rui Li, Thibaud Frere, Leandro von Werra, Aritra~Roy Gosthipaty, and Andr{\'e}s Marafioti. 2025.
\newblock Finevision: Open data is all you need.
\newblock \emph{arXiv preprint arXiv:2510.17269}.

\bibitem[{Yang et~al.(2025)Yang, Zhu, Lu, Wang, Chen, Gao, Yan, and Chen}]{SurveyKD}
Chuanpeng Yang, Yao Zhu, Wang Lu, Yidong Wang, Qian Chen, Chenlong Gao, Bingjie Yan, and Yiqiang Chen. 2025.
\newblock \href {https://doi.org/10.1145/3699518} {Survey on knowledge distillation for large language models: Methods, evaluation, and application}.
\newblock \emph{ACM Trans. Intell. Syst. Technol.}, 16(6).

\bibitem[{Yu et~al.(2024)Yu, Jiang, Shi, YU, Liu, Zhang, Kwok, Li, Weller, and Liu}]{yu2024metamath}
Longhui Yu, Weisen Jiang, Han Shi, Jincheng YU, Zhengying Liu, Yu~Zhang, James Kwok, Zhenguo Li, Adrian Weller, and Weiyang Liu. 2024.
\newblock \href {https://openreview.net/forum?id=N8N0hgNDRt} {Metamath: Bootstrap your own mathematical questions for large language models}.
\newblock In \emph{The Twelfth International Conference on Learning Representations}.

\bibitem[{Yue et~al.(2024{\natexlab{a}})Yue, Ni, Zhang, Zheng, Liu, Zhang, Stevens, Jiang, Ren, Sun et~al.}]{mmmu}
Xiang Yue, Yuansheng Ni, Kai Zhang, Tianyu Zheng, Ruoqi Liu, Ge~Zhang, Samuel Stevens, Dongfu Jiang, Weiming Ren, Yuxuan Sun, et~al. 2024{\natexlab{a}}.
\newblock Mmmu: A massive multi-discipline multimodal understanding and reasoning benchmark for expert agi.
\newblock In \emph{Proceedings of the IEEE/CVF Conference on Computer Vision and Pattern Recognition}, pages 9556--9567.

\bibitem[{Yue et~al.(2024{\natexlab{b}})Yue, Zheng, Ni, Wang, Zhang, Tong, Sun, Yu, Zhang, Sun, Su, Chen, and Neubig}]{mmmupro}
Xiang Yue, Tianyu Zheng, Yuansheng Ni, Yubo Wang, Kai Zhang, Shengbang Tong, Yuxuan Sun, Botao Yu, Ge~Zhang, Huan Sun, Yu~Su, Wenhu Chen, and Graham Neubig. 2024{\natexlab{b}}.
\newblock Mmmu-pro: A more robust multi-discipline multimodal understanding benchmark.
\newblock \emph{arXiv preprint arXiv:2409.02813}.

\bibitem[{Zhang et~al.(2025{\natexlab{a}})Zhang, Wu, Yang, Li, Hu, Wang, Liu, Li, and Bing}]{openmmreasoner}
Kaichen Zhang, Keming Wu, Zuhao Yang, Bo~Li, Kairui Hu, Bin Wang, Ziwei Liu, Xingxuan Li, and Lidong Bing. 2025{\natexlab{a}}.
\newblock Openmmreasoner: Pushing the frontiers for multimodal reasoning with an open and general recipe.
\newblock \emph{arXiv preprint arXiv:2511.16334}.

\bibitem[{Zhang et~al.(2024)Zhang, Jiang, Zhang, Lin, Guo, Qiu, Zhou, Lu, Chang, Qiao, Gao, and Li}]{mathverse}
Renrui Zhang, Dongzhi Jiang, Yichi Zhang, Haokun Lin, Ziyu Guo, Pengshuo Qiu, Aojun Zhou, Pan Lu, Kai-Wei Chang, Yu~Qiao, Peng Gao, and Hongsheng Li. 2024.
\newblock \href {https://doi.org/10.1007/978-3-031-73242-3_10} {Mathverse: Does your multi-modal llm truly see the diagrams in visual math problems?}
\newblock In \emph{Computer Vision -- ECCV 2024: 18th European Conference, Milan, Italy, September 29--October 4, 2024, Proceedings, Part VIII}, pages 169--186, Berlin, Heidelberg. Springer-Verlag.

\bibitem[{Zhang et~al.(2025{\natexlab{b}})Zhang, Li, Long, Zhang, Lin, Yang, Xie, Yang, Liu, Lin et~al.}]{qwen3-embedding}
Yanzhao Zhang, Mingxin Li, Dingkun Long, Xin Zhang, Huan Lin, Baosong Yang, Pengjun Xie, An~Yang, Dayiheng Liu, Junyang Lin, et~al. 2025{\natexlab{b}}.
\newblock Qwen3 embedding: Advancing text embedding and reranking through foundation models.
\newblock \emph{arXiv preprint arXiv:2506.05176}.

\bibitem[{Zhang et~al.(2026)Zhang, Ni, Chen, Zhang, Rao, Peng, Lu, Hu, Guo, and min Hu}]{bee}
Yi~Zhang, Bolin Ni, Xin-Sheng Chen, Hengrui Zhang, Yongming Rao, Houwen Peng, Qinglin Lu, Han Hu, Meng-Hao Guo, and Shi min Hu. 2026.
\newblock \href {https://openreview.net/forum?id=IVluwK8q9q} {Bee: A high-quality corpus and full-stack suite to unlock advanced fully open {MLLM}s}.
\newblock In \emph{The Fourteenth International Conference on Learning Representations}.

\bibitem[{Zheng et~al.(2023)Zheng, Chiang, Sheng, Zhuang, Wu, Zhuang, Lin, Li, Li, Xing, Zhang, Gonzalez, and Stoica}]{Chatbot-Arena}
Lianmin Zheng, Wei-Lin Chiang, Ying Sheng, Siyuan Zhuang, Zhanghao Wu, Yonghao Zhuang, Zi~Lin, Zhuohan Li, Dacheng Li, Eric~P. Xing, Hao Zhang, Joseph~E. Gonzalez, and Ion Stoica. 2023.
\newblock Judging llm-as-a-judge with mt-bench and chatbot arena.
\newblock In \emph{Proceedings of the 37th International Conference on Neural Information Processing Systems}, NIPS '23, Red Hook, NY, USA. Curran Associates Inc.

\end{thebibliography}
\end{document}